\pdfoutput=1
\documentclass{interact}
\usepackage{amsmath,amssymb,amsthm}
\usepackage{graphicx}
\usepackage[T1]{fontenc}
\usepackage{lmodern}
\usepackage{booktabs,tabularx}
\usepackage[round,authoryear]{natbib}
\usepackage{xcolor}
\usepackage[
  colorlinks=false,
  pdfborder={0 0 1},
  citebordercolor={0 1 0},
  linkbordercolor={0 1 0},
  urlbordercolor={0 1 0}
]{hyperref}
\usepackage{adjustbox}
\usepackage[section]{placeins}

\theoremstyle{definition}

\newcommand{\R}{\mathbb{R}}
\graphicspath{{figures/}}
\newcommand{\bp}{\mathbf{p}}
\newcommand{\bdelta}{\boldsymbol{\delta}}
\newcommand{\bu}{\mathbf{u}}
\newcommand{\bx}{\mathbf{x}}
\newcommand{\bh}{\mathbf{h}}
\newcommand{\bphi}{\boldsymbol{\phi}}
\newcommand{\bJ}{\mathbf{J}}
\newcommand{\bG}{\mathbf{G}}
\newcommand{\bB}{\mathbf{B}}

\title{Tissot indicatrix for location encoders: Characterizing local distortion in learned geographic embeddings}

\author{\name{Guofeng Cao\textsuperscript{a}\thanks{CONTACT Guofeng Cao. Email: guofeng.cao@colorado.edu} and Gengchen Mai\textsuperscript{b}}\affil{\textsuperscript{a}Department of Geography, University of Colorado Boulder, Boulder, CO, USA; \textsuperscript{b}Department of Geography and the Environment, University of Texas at Austin, Austin, TX, USA}}

\begin{document}

\maketitle

\begin{abstract}
\noindent With the increasing adoption of geographic artificial intelligence, location encoders are widely used to map geographic coordinates to learned embeddings for tasks in Earth observation and geospatial modeling.
Similar to map projections, location encoders can distort geographic space, that is, equal-distance displacements in different directions can result in unequal embedding changes.
However, such distortion remains largely unexamined, and it can propagate into downstream applications.
We adapt Tissot's indicatrix, a cartographic tool for map projection distortion, to characterize this distortion.
At each location, we compare encoder outputs at equal distances in 32 directions and fit a tensor, represented by an ellipse, whose elongation (anisotropy) and orientation describe the local distortion.
The fit requires only encoder outputs without access to model internals and is validated against references with known geometry.
By applying the method to thirteen released checkpoints of five encoder families, we find that distortions vary across space and inherit the input geometry to varying degrees, closely in GeoCLIP with its Equal Earth input (anisotropy correlation of 0.88) and weakly or moderately in the other encoders.
The distortion also affects the locations an encoder treats as neighbors.
Within 1\,km of $20$ query sites, the $40$ nearest neighbors in geographic and embedding space overlap by a median of only 57.5--87.5\% across the checkpoints, while the fitted ellipses select exactly the same embedding neighbors in $265$ of $299$ comparisons.
These findings provide a geographic basis for assessing location encoders, and the accompanying \texttt{eTissot} package implements the methods for user-specified encoders.
\end{abstract}

\begin{keywords}\textit{Geographic artificial intelligence; location embeddings; embedding geometry; Tissot indicatrix; map projection; Earth observation}\end{keywords}

\section{Introduction}\label{sec:intro}

In geographic artificial intelligence (GeoAI), location encoders have emerged as widely adopted spatial representation learning models \citep{mai2024srl} that map geographic coordinates to learned high-dimensional embeddings. They allow geographic locations to be compared in feature space and can be directly used in downstream applications.
These encoders are usually trained in a supervised or self-supervised manner with a variety of georeferenced data (e.g., satellite imagery, geotagged photos, species observations and climate data) \citep{mai2022review,klemmer2024satclip,cepeda2023geoclip,cole2023sinr,dollinger2025climplicit,liu2026gair}. 
Distances or similarity between these learned embeddings inherit their spatial relation in the original geographic space and can be utilized in spatial prediction and retrieval tasks.
However, these distances in the embedding space do not necessarily correspond to geographic distance, i.e., the local isometry \citep{gao2018learning} might not be preserved in the learned location embedding space. 
Similar to a map projection, an encoder can translate equal-distance displacements in different directions into unequal changes in its embeddings, and such directional response can vary from place to place.
We refer to this unequal response as \textit{local distortion} for embeddings. 
It can affect the locations an encoder treats as neighbors: given a query location, a geographically distant location along a direction of slow embedding change can have a more similar embedding than a nearby location along a direction of rapid change.
This is of particular relevance to Earth observation (EO), where location embeddings are commonly used as input features or geographic priors in prediction models \citep{macaodha2019,mai2023csp,klemmer2024satclip}, so that the local distortion of the embeddings can propagate into their predictions.
 
In cartography, Tissot's indicatrix provides a framework for characterizing local distortion in map projections. 
It describes the distortion as an ellipse into which the projection deforms an infinitesimal circle on the Earth's surface \citep{tissot1881,snyder1987}, whose semiaxes measure the greatest and least stretching and associated directions.
Distortion indices, finite-distance errors, and higher-order local measures (e.g., flexion and skewness) extend the analysis beyond a single infinitesimal ellipse \citep{tobler1963,milnor1969,laskowski1997,canters2002,gott2007distance,goldberg2007}.
A location encoder has many more output dimensions than a map projection, but its input still varies in only two local directions. 
Its local linear approximation therefore also maps a small circle on the Earth's surface to an ellipse, whose semiaxes give the greatest and least stretching of the encoder at that location.

Geographic representation learning has already addressed the distortion of spherical distance and projection in the construction of positional encodings.
Sphere2Vec, for example, uses Tissot indicatrices to illustrate distortion in input map projections \citep{mai2023sphere2vec}, and 
proposes a unified view of spherical-distance-preserving position encoding based on Double Fourier Sphere. 
SatCLIP also utilizes spherical-harmonic encodings to incorporate the geometry of the sphere into the input representation \citep{russwurm2024satclip}.
These studies focus on the input stage of the location encoders, 
 i.e., the position encodings, while the embedding geometry obtained after a learned transformation remains largely unexamined.
Recently, \citet{rao2026intrinsic} mapped the local intrinsic dimension (the number of degrees of freedom in local variations) of several encoders and reported spatial artifacts such as grid patterns and harmonic oscillations, but did not measure how much, or in which directions, the encoder stretches geographic space.

More generally, the pullback metric (or first fundamental form in surface geometry; \citealp{docarmo1976}) measures the geometric distortion that a function (e.g., a neural network) induces on its input space, and it has been used to study the distortions of learned representations \citep{hauser2017riemannian,arvanitidis2018,zavatoneveth2026}.
Indicatrices of the pullback metric have been used to visualize the local geometry of autoencoders with two latent dimensions \citep{nazari2023geometric}; in this terminology, the reach ellipse developed in this paper (Section~\ref{sec:setting}) is the indicatrix of the pullback metric of a location encoder, scaled by a feature distance.
Different from these general settings, the input of location encoders lies in geographic space, with known coordinates and distances.
Hence, the embedding change corresponding to a geographic displacement can be characterized per unit of ground distance in each direction, which allows the local distortion and its spatial pattern to be measured across the globe. 
For differentiable encoders, the model Jacobian \citep{docarmo1976}, computed by automatic differentiation \citep{baydin2018autodiff}, gives this change directly for an infinitesimal displacement, and, together with positional encodings \citep{mai2023sphere2vec,russwurm2024satclip} and map projections \citep{snyder1987,savric2019} whose geometry is known in closed form, it provides a reference for assessing the accuracy of the measurement.
To our knowledge, the local distortion of location encoders, including the anisotropy and orientation fields of released encoders and their implications for selecting spatial neighborhoods, has not been characterized.

To address this gap, this paper makes three contributions to the assessment of location encoders. 
First, we develop a protocol to measure the local distortion of an encoder and to assess the accuracy of the measurement. 
At each location, we evaluate the encoder with an equal-length geodesic probe along equally spaced bearings, and fit a harmonic least-squares estimator to summarize its directional response as a Tissot-style ellipse in geographic distances and bearings. 
The fit requires only the encoder outputs at sampled locations without access to its internal implementation, so it also applies to precomputed embedding products that are available at the probe locations.
We assess the accuracy of the fit against closed-form references and model Jacobians when available, and find close agreement except where ReLU activations make the encoder non-smooth.
Second, we apply the protocol to thirteen released checkpoints from five commonly used encoder families (Table~\ref{tab:encoders}) to characterize the spatial variation of their local distortion across the globe and its relationship to their input encodings, and find that only GeoCLIP closely follows the geometry of its input encoding.
Third, we examine how local distortion affects the spatial neighborhoods defined by embedding similarity. 
By comparing geographic neighbors, embedding neighbors, and the neighbors predicted by the fitted ellipse, we test whether the measured geometry predicts departures from geographic proximity, and find that the fitted ellipse reproduces the embedding neighbors in most comparisons. 
We use reach ellipses (Section~\ref{sec:setting}) to show how the extent of an embedding neighborhood varies with direction.
These methods are implemented in the accompanying \texttt{eTissot} Python package, which can be applied to user-supplied encoders.

The remainder of this paper is organized as follows.
Section~\ref{sec:background} reviews map projection distortion and location encoders, and Section~\ref{sec:theory} introduces the reach ellipse, a Tissot-style indicatrix, and its estimation.
Section~\ref{sec:validation} assesses the estimator against closed-form references and model Jacobians, Section~\ref{sec:encoders} characterizes the local geometry of the released encoders and its effect on neighborhoods, and Section~\ref{sec:discussion} discusses implications and limitations before Section~\ref{sec:conclusion} concludes.

\section{Background}\label{sec:background}

\subsection{Cartographic and representation geometry}\label{sec:cartography}

For a smooth map projection with two positive principal stretches, its local linear approximation maps an infinitesimal circle on the Earth's surface to an ellipse.
The long and short semiaxes, $a\ge b$, measure the greatest and least stretch, their ratio $a/b$ measures \textit{anisotropy} (or directional inequality) \citep{snyder1987}, and the direction of the long axis gives the \textit{orientation} of greatest stretch.
Tissot's indicatrix also gives the maximum angular deformation, $\omega=2\arcsin\{(a-b)/(a+b)\}$, and the areal scale, $ab$, which measure how much a projection distorts angles and areas \citep{snyder1987}.
These quantities have been used by cartographers to assess and design projections \citep{jenny2008,savric2019}.

Similar constructions have been developed to characterize the local geometry of embedding spaces of higher dimension.  
Studies of distance-preserving embeddings and representation geometry \citep{verma2013,arvanitidis2018}, local metric estimation in manifold learning \citep{perraultjoncas2013,meila2024} and Tissot-style displays of dimension reduction outputs \citep{sankaran2026} provide examples of such developments. 
These studies focus on abstract input spaces, such as the latent spaces of generative models, without a standard reference for distance or direction.
Location encoders, however, have a geographic input space with known coordinates and distances, which allows the characterization of local distortion in geographic terms, as developed in Section~\ref{sec:theory}.

\subsection{Location encoders}

Location encoders transform geographic coordinates into high-dimensional feature vectors that can be used for downstream analysis tasks \citep{mai2022review}.
The construction of a location encoder involves two related components: how coordinates are represented and what information the model learns from georeferenced observations.
Typically, an encoder first transforms coordinates into numerical input features through a positional encoding $\bphi$, and then applies a learned transformation often represented by a neural network $\bh$, which gives the outputs $\mathbf{f}=\bh\circ\bphi$. 
Separating the two components distinguishes the spatial structure specified at the input stage from the structure learned from training data.

Positional encodings provide various ways to represent variation at different spatial scales and to account for the geometry of the geographic domain.
For example, Space2Vec uses multiscale sine and cosine functions to represent planar locations \citep{mai2020space2vec}, whereas Sphere2Vec constructs multiscale encodings on the sphere and derives their distance-preservation properties \citep{mai2023sphere2vec}.
Spherical harmonics combined with sinusoidal networks provide another approach to representing locations on the Earth's surface \citep{russwurm2024satclip}, while GeoCLIP applies random Fourier features at multiple scales to coordinates in the Equal Earth projection \citep{savric2019,tancik2020fourier,cepeda2023geoclip}.
These choices determine the representation of coordinates before entering a neural network, which can subsequently change the representation geometry.

Training connects these coordinate representations to the information associated with locations.
In the manner of CLIP (contrastive language-image pre-training), GeoCLIP and CSP (contrastive spatial pre-training) use contrastive learning to align location and image representations, encouraging corresponding pairs to be more similar than unrelated pairs \citep{cepeda2023geoclip,mai2023csp}, and SatCLIP uses satellite imagery for this alignment \citep{klemmer2024satclip}.
Other encoders learn through prediction, such as SINR (spatial implicit neural representations), which predicts species occurrence from location, and Climplicit, which learns representations of climate \citep{cole2023sinr,dollinger2025climplicit}.
The five families examined here therefore differ in both their input constructions and their training information, and Table~\ref{tab:encoders} lists the positional encoding $\bphi$, the network $\bh$ and the training data of each family.
Their learned transformations also differ in smoothness, which is important for local measurement.
Specifically, the sinusoidal representation networks \citep[SIREN;][]{sitzmann2020siren} of SatCLIP and Climplicit and the Gaussian error linear unit \citep[GELU;][]{hendrycks2016gelu} activations of CSP-fMoW are smooth (infinitely differentiable), whereas the rectified linear unit (ReLU) or leaky ReLU activations of GeoCLIP, SINR and CSP-iNat make them piecewise linear (i.e., smooth within each activation region but with abrupt changes of derivative across region boundaries), which affects how closely a finite-distance fit approximates the model Jacobian (Section~\ref{sec:jacobian}).
For each released checkpoint (i.e., saved set of trained weights), the local geometry of a specified feature vector (Section~\ref{sec:protocol}) is measured and compared with that of its positional encoding, the input reference (Section~\ref{sec:protocol}).

\begin{table}[htbp]
\centering\small
\caption{Location encoders examined in this paper. $\lambda$ and $\varphi$ denote longitude and latitude; network widths and output dimensions are those of the released checkpoints (provenance in supplementary Section~\ref{supp:checkpoints}).}
\label{tab:encoders}
\begin{tabularx}{\textwidth}{@{}>{\raggedright\arraybackslash}p{2.9cm}>{\raggedright\arraybackslash}X>{\raggedright\arraybackslash}X>{\raggedright\arraybackslash}p{2.6cm}@{}}
\toprule
Family (checkpoints) & Positional encoding $\bphi$ & Learnable network $\bh$ (output dimension) & Training data\\\midrule GeoCLIP (1)\newline\citealp{cepeda2023geoclip} & Equal Earth coordinates with random Fourier features at three scales & Three ReLU multilayer perceptrons (three hidden layers of 1,024 units), outputs summed (512) & Geotagged photographs, contrastive\\[4pt]
SatCLIP (6)\newline\citealp{klemmer2024satclip} & Spherical harmonics of degree $l<L$, $L=10$ or 40 & SIREN with two hidden layers of 512 units (256) & Sentinel-2 images, contrastive; ViT, ResNet-18 or ResNet-50 image encoder\\[4pt]
SINR (3)\newline\citealp{cole2023sinr} & $(\sin\lambda,\cos\lambda,\sin2\varphi,\cos2\varphi)$ & ReLU residual network with four residual blocks of 256 units (256, before the species output layer) & Species presence records, prediction\\[4pt]
CSP (2)\newline\citealp{mai2023csp} & Space2Vec grid: sinusoids of $\lambda$ and $\varphi$ at 32 scales with equal frequencies for both & One hidden layer of 512 units with layer normalization; leaky ReLU (iNat) or GELU (fMoW) (256) & iNaturalist 2018 or fMoW images, contrastive\\[4pt]
Climplicit (1)\newline\citealp{dollinger2025climplicit} & $(\lambda/\pi,\,2\varphi/\pi)$ with the month as sine and cosine & SIREN with 16 layers of 512 units, residual connections and a hyperbolic-sine first layer (256 per month) & Climate data, prediction\\
\bottomrule
\end{tabularx}
\end{table}

\subsection{Evaluation of location encoders}\label{sec:evaluation}

Existing studies have assessed geographic representations from different perspectives, including predictive performance and geographic bias \citep{wu2024torchspatial, liu2026gair}, recovery of known spatial effects \citep{kiv2026geoshapley}, preservation of topology, direction and distance \citep{siampou2025poly2vec}, generalization across geographic distributions \citep{crasto2025robustness}, interpretable content \citep{betti2026earth} and regional resolution \citep{rao2026slepian}.
Our analysis complements these assessments by characterizing the distortion of a specified encoder checkpoint responding to geographic displacements in different directions, and related effects on neighborhood selection (Section~\ref{sec:neighborhoods}).

\section{A geographic indicatrix}\label{sec:theory}

Building on the cartographic construction of Section~\ref{sec:cartography}, we characterize local distortion by comparing an encoder's response to equal-distance geographic displacements at different bearings.
The infinitesimal response defines a local Tissot ellipse, which we estimate from feature differences around a finite-radius ring, and repeating the measurement across locations produces a geographic field of anisotropy and orientation.
Section~\ref{sec:setting} defines the local metric and the reach ellipse, Section~\ref{sec:finite} estimates them from finite-radius probes, and Section~\ref{sec:validity} discusses accuracy and the choice of radius.

\subsection{Local metric and reach ellipse}\label{sec:setting}

Let $\mathbf{f}:U\to\R^d$ be a location encoder that maps each location $\bp = (\lambda, \varphi)$ in the geographic domain $U$ to a feature vector with $d$ components, where $\lambda$ and $\varphi$ denote the longitude and latitude at a location $\bp$.
We represent the Earth by a sphere of radius $R=6371$\,km in the following discussion.
At each location, we describe displacements by their northward and eastward components in kilometers, which form a local north-east frame.
For a differentiable encoder $\mathbf{f}(\cdot)$, its response over a sufficiently small geographic displacement $\bdelta$ can be approximated by a linear transformation according to Taylor expansion:
\begin{equation} \label{eq:taylor}
\mathbf{f}(\bp+\bdelta)=\mathbf{f}(\bp)+\bJ_{\bp}\bdelta+o(\|\bdelta\|_2),
\end{equation}
where $\bp+\bdelta$ is the point reached by traveling $\|\bdelta\|_2$ along the great circle in the direction of $\bdelta$, and $\bJ_{\bp}$ is the $d\times2$ Jacobian matrix of $\mathbf{f}$ at $\bp$.
In the north-east frame, the Jacobian matrix can be written as
\begin{equation}\label{eq:jacobian}
\bJ_{\bp}=\left[\,\frac{1}{R}\frac{\partial \mathbf{f}}{\partial\varphi}\quad\frac{1}{R\cos\varphi}\frac{\partial \mathbf{f}}{\partial\lambda}\,\right]\in\R^{d\times2},
\end{equation}
whose columns give the rates of change of each feature component with northward and eastward displacement, since $R$ and $R\cos\varphi$ are the ground distances spanned by one radian of latitude $\varphi$ and longitude $\lambda$. Hence the distance corresponding to $\bdelta$ in embedding space can be expressed as:
\begin{equation}\label{eq:quadratic-form}
\|\mathbf{f}(\bp+\bdelta)-\mathbf{f}(\bp)\|_2^2=\bdelta^\top \bG(\bp)\bdelta+o(\|\bdelta\|_2^2),\qquad \bG(\bp)=\bJ_{\bp}^\top \bJ_{\bp},
\end{equation}
where $\bG(\bp)$ is the pullback metric (or local metric tensor) of $\mathbf{f}$, which describes the geometric distortion induced by $\mathbf{f}$.
Because geographic displacement has only two local directions, this matrix is $2\times2$ regardless of the feature dimension $d$, with diagonal entries giving the squared stretch northward and eastward, and the off-diagonal entry describing how the two combine for oblique displacement.
The eigenvalues $\lambda_1\ge\lambda_2$ of $\bG(\bp)$ are the greatest and least squared stretches, and the eigenvector associated with $\lambda_1$ gives the principal bearing (i.e., the direction of greatest stretch), defined modulo $180^\circ$.
The anisotropy is the ratio of the greatest to the least stretch, corresponding to $a/b$ for a map projection:
\begin{equation}\label{eq:metric-anisotropy}
A(\bp)=\sqrt{\lambda_1/\lambda_2},\quad \lambda_1\ge\lambda_2>0.
\end{equation}
Similar to map projections, anisotropy $A=1$ means equal stretch in all directions, where the principal bearing is undefined; $A=2$ means twice as much stretch along the principal bearing as across it.
Anisotropy and bearing are invariant to a uniform scaling, translation or rotation applied to all feature vectors, and because $\bG(\bp)$ is expressed in north-east geographic distances, they are comparable across encoders once the measured feature vector is specified (e.g., normalized vs. raw vectors; see Section~\ref{sec:protocol}).
For the other measures in Section~\ref{sec:cartography}, the maximum angular deformation is a function of $A$, $\omega=2\arcsin\{(A-1)/(A+1)\}$, and the areal scale, $\sqrt{\lambda_1\lambda_2}$, carries the units of the feature space, which differ across encoders and change with normalization. We therefore focus on anisotropy and principal bearing for assessing and comparing encoders.

\begin{figure}[htbp]
\centering\includegraphics[width=\textwidth]{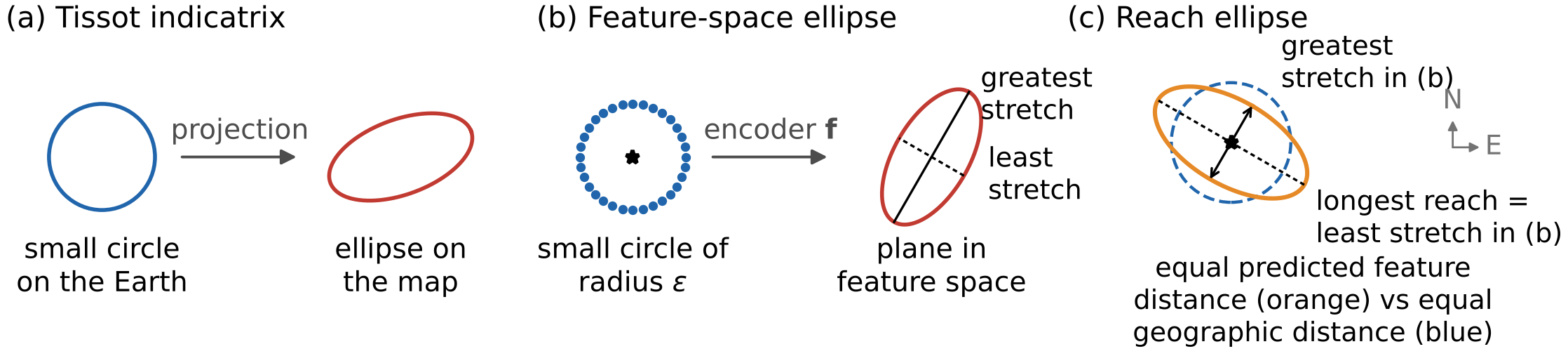}
\caption{The three ellipses used in this paper, for $A=2$: (a) Tissot’s indicatrix of a map projection (Section~\ref{sec:cartography}); (b) the ellipse in feature space which the linear approximation of the encoder maps a small ground circle to, with axes along the greatest and least stretch (Section~\ref{sec:setting}); (c) the reach ellipse (Equation~\ref{eq:reach}; orange), with the dashed circle marking equal geographic distance.}
\label{fig:schematic}
\end{figure}

Similar to Tissot's indicatrix (Figure~\ref{fig:schematic}a) that represents the distortion of a map projection with an ellipse which an infinitesimal ground circle is mapped to, the local geometric distortion of an encoder can be described by an ellipse determined by the tensor $\bG(\bp)$.
Specifically, given a small ground circle $\{\epsilon\bu(\theta):0\le\theta<2\pi\}$, where $\epsilon>0$ is a small radius and $\bu(\theta)=(\cos\theta,\sin\theta)^\top$ is the unit direction at bearing $\theta$, the linear approximation of the encoder maps this circle to the ellipse $\{\epsilon\bJ_{\bp}\bu(\theta):0\le\theta<2\pi\}$ in the plane of feature space spanned by the two columns of $\bJ_{\bp}$ (Equation~\ref{eq:taylor}; Figure~\ref{fig:schematic}b). 
Since this ellipse lies in feature space and has no geographic orientation, we represent the local distortion on the ground by the \textit{reach ellipse} (Figure~\ref{fig:schematic}c). 
Given a location at $\bp$, the reach ellipse represents the set of ground displacements from $\bp$ sharing the same feature distance $c>0$, i.e.,
\begin{equation}\label{eq:reach}
\{\bdelta:\bdelta^\top \bG(\bp)\bdelta=c^2\}.
\end{equation}
It has semiaxes $c/\sqrt{\lambda_2}\ge c/\sqrt{\lambda_1}$, so its axis ratio is the anisotropy $A(\bp)$, and its long axis is perpendicular to the principal bearing of $\bG(\bp)$, since a displacement reaches farther where the features change more slowly.

\subsection{Directional measurement and tensor fitting}\label{sec:finite}

To estimate $\bG(\bp)$ and associated ellipses from encoder outputs alone, we evaluate the encoder at a fixed geographic distance from $\bp$ in equally spaced directions and fit the tensor to the resulting directional profile of feature change.
Let $\bp_r(\theta)$ be the point reached from $\bp$ by traveling a distance $r$ along a great-circle path at bearing $\theta$, measured clockwise from north. The finite directional stretch is
\begin{equation}\label{eq:finite-stretch}
s_r(\bp,\theta)=\frac{\|\mathbf{f}(\bp_r(\theta))-\mathbf{f}(\bp)\|_2}{r},
\end{equation}
where $\|\cdot\|_2$ is the Euclidean norm over the selected feature vector (Section~\ref{sec:protocol}), so that $s_r$ measures feature change per kilometer of geographic distance change. 
In the notation of Section~\ref{sec:setting}, $\bp_r(\theta)=\bp+r\bu(\theta)$, so Equation~\ref{eq:quadratic-form} with $\bdelta=r\bu(\theta)$ gives $s_r^2(\bp,\theta)=\bu(\theta)^\top \bG(\bp)\bu(\theta)+o(1)$. 
As $r\to0$, the finite stretch converges to the limiting stretch $s_0(\bp,\theta)$, with $\bp$ omitted for brevity:
\begin{equation}\label{eq:limiting-stretch}
s_0^2(\theta)=\lim_{r\to0}s_r^2(\theta)=\bu(\theta)^\top \bG\bu(\theta)=a_0+a_2\cos2\theta+b_2\sin2\theta,
\end{equation}
with $a_0=(G_{NN}+G_{EE})/2$, $a_2=(G_{NN}-G_{EE})/2$ and $b_2=G_{NE}$; subscripts $N,E$ denote north and east (see supplementary Section~\ref{supp:harmonic} for the derivation details).
The coefficients describe the mean squared stretch, the north-east difference, and the contribution of oblique directions, respectively. Note that we focus on the squared stretch because it allows the tensor to be estimated by a linear fit.

For an infinitesimal radius, the directional stretch is symmetric about opposite directions ($\bu(\theta+\pi)=-\bu(\theta)$), hence the limiting stretch $s_0^2(\theta)$ (Equation~\ref{eq:limiting-stretch}) contains only constant and double-angle terms. 
This is not the case for a finite radius ($s_r^2(\theta)$), however, as the encoder can respond differently to opposite directions. 
We therefore add the first-harmonic terms to account for this asymmetry:
\begin{equation}\label{eq:profile-fit}
s_r^2(\theta_j)\approx a_{0,r}+a_{2,r}\cos2\theta_j+b_{2,r}\sin2\theta_j
+a_{1,r}\cos\theta_j+b_{1,r}\sin\theta_j
\end{equation}
\noindent where the constant and double-angle terms describe the tensor part (as in Equation~\ref{eq:limiting-stretch}), and the terms with coefficients $a_{1,r},b_{1,r}$ measure this asymmetry.
Given a location $\bp$, we estimate this directional model from $n=32$ equally spaced bearings, $\theta_j=2\pi j/n$ for $j=0,\ldots,n-1$. 
Based on the measured squared stretches $y_j=s_r^2(\bp,\theta_j)$, we fit the model (Equation \ref{eq:profile-fit}) by ordinary least squares, giving the estimates $\widehat a_{0,r},\ldots,\widehat b_{1,r}$. 
Note that we fit separate models in different locations.

\begin{figure}[htbp]
\centering\includegraphics[width=\textwidth]{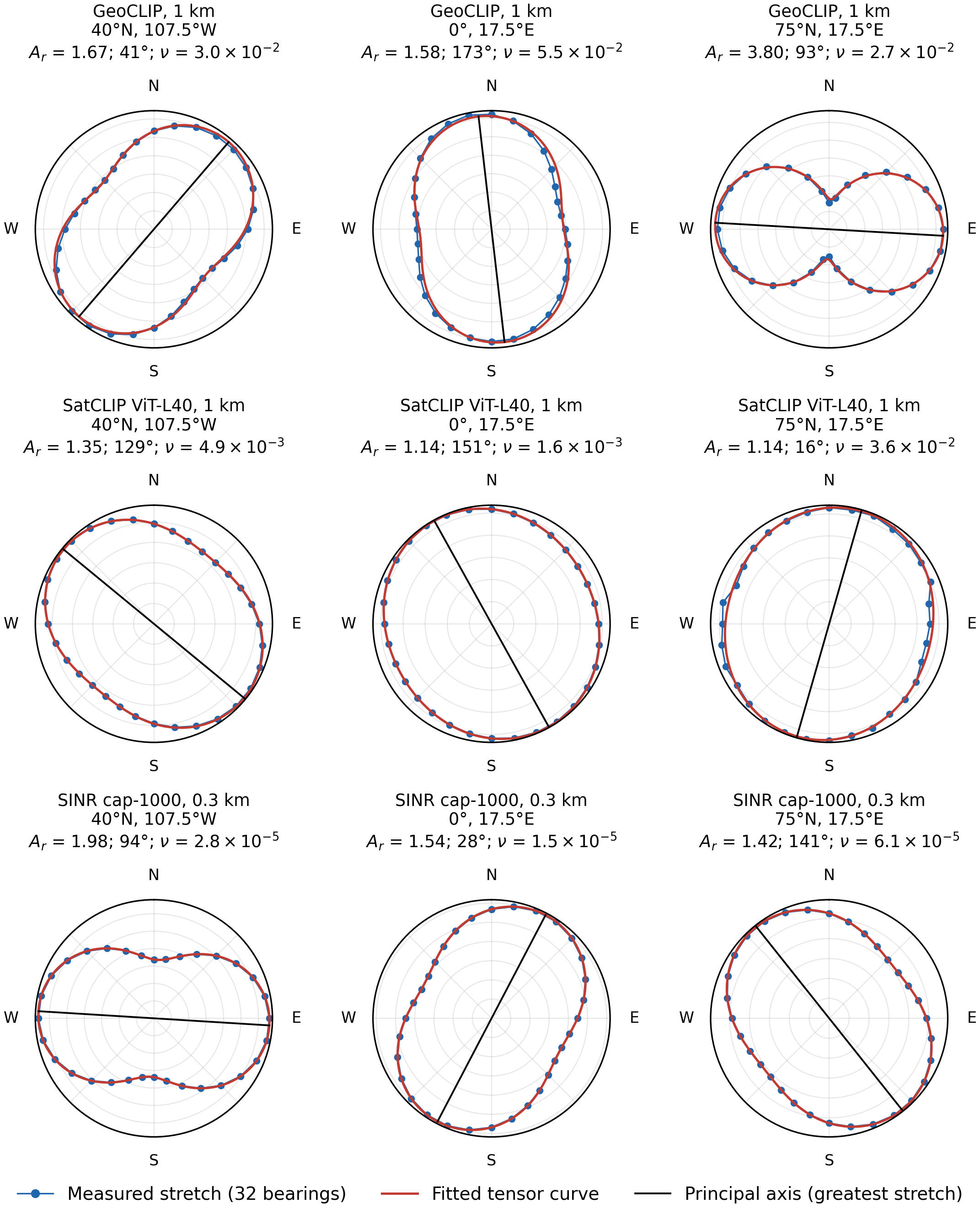}

\caption{Directional profiles of GeoCLIP, SatCLIP ViT-L40 and SINR cap-1000 (rows), each at its primary radius, at three sites (columns). In each polar plot, the distance from the center gives the stretch at a bearing measured clockwise from north: blue dots show the measured stretch at the 32 probe bearings (Equation~\ref{eq:finite-stretch}), the red curve shows the estimated limiting stretch $\sqrt{\widehat{s_0^2}(\theta)}$ (Section~\ref{sec:finite}), and the black line marks the principal axis. Stretches are normalized by their geometric mean in each panel, and panel titles give $A_r$, the principal bearing (modulo $180^\circ$) and $\nu$ (Equation~\ref{eq:nonellipticity}). Note that the stretch is measured in feature space rather than in ground distance.}

\label{fig:indicatrix}
\end{figure}

Since the first-harmonic terms cannot be represented by a tensor and vanish as $r\to0$ (supplementary Section~\ref{supp:harmonic}), we build the fitted tensor for a finite radius $r$, the counterpart of $\bG(\bp)$, from the constant and double-angle coefficients only.
As the constant and double-angle terms are orthogonal to the first-harmonic terms on equally spaced bearings, their estimated coefficients are not affected by the asymmetry \citep{bloomfield2000}, and the fitted tensor is:

\begin{equation}\label{eq:fitted-tensor}
\widehat{\bG}_r=\begin{pmatrix}
\widehat a_{0,r}+\widehat a_{2,r}&\widehat b_{2,r}\\
\widehat b_{2,r}&\widehat a_{0,r}-\widehat a_{2,r}
\end{pmatrix}.
\end{equation}
\noindent When its ordered eigenvalues $\widehat\lambda_{1,r},\widehat\lambda_{2,r}$ are positive, they give the fitted anisotropy $A_r=\sqrt{\widehat\lambda_{1,r}/\widehat\lambda_{2,r}}$ (the finite-radius counterpart of Equation~\ref{eq:metric-anisotropy}), and if they are distinct, the principal eigenvector gives the bearing $\theta_r$.
The tensor part of the fitted profile (Equation~\ref{eq:profile-fit}), $\widehat{s_0^2}(\theta)=\bu(\theta)^\top\widehat{\bG}_r\bu(\theta)=\widehat a_{0,r}+\widehat a_{2,r}\cos2\theta+\widehat b_{2,r}\sin2\theta$, therefore estimates the limiting squared stretch $s_0^2(\theta)$ of Equation~\ref{eq:limiting-stretch}.
As an illustration, Figure~\ref{fig:indicatrix} shows the measured directional profiles and this estimate at sample locations.

We assess how closely the tensor reproduces the measured profile using the non-ellipticity $\nu$, defined as the root-mean-square (RMS) difference between the measured squared stretches $y_j$ and their estimated tensor part $\widehat{s_0^2}(\theta_j)$, normalized by the fitted mean:
\begin{equation}\label{eq:nonellipticity}
\nu=\frac{\sqrt{n^{-1}\sum_{j=0}^{n-1}
[y_j-\widehat{s_0^2}(\theta_j)]^2}}
{\widehat a_{0,r}}.
\end{equation}
The denominator $\widehat a_{0,r}$, the estimated mean squared stretch, has the same units as the residuals in the numerator, so $\nu$ is dimensionless and invariant to a uniform rescaling of the outputs.
The residuals also include the asymmetry between opposite directions, which accounts for most of the mean squared residual in $\nu$ at every checkpoint but does not affect the fitted tensor (supplementary Section~\ref{supp:harmonic}).
Because the mean squared stretch is dominated by the greatest stretch, $\nu$ can be small even when the least stretch is poorly fitted, and it therefore does not bound anisotropy error (see supplementary Section~\ref{supp:bound} for an example).
Since $A_r$ is inversely proportional to the square root of the fitted least squared stretch $\widehat\lambda_{2,r}$, a non-elliptic profile can exaggerate anisotropy through a fitted value far below the sampled values.
In those cases, a flagged fallback replaces it with the smallest sampled squared stretch; such fallbacks occur at fewer than 1\% of global-grid sites for every checkpoint (supplementary Section~\ref{supp:harmonic}).
For the same reason, the anisotropy error tends to be larger at strongly anisotropic locations where the least squared stretch is small (Section~\ref{sec:jacobian}), while the bearing becomes unstable where the two principal stretches are nearly equal.

\subsection{Relation to the Jacobian and choice of probe radius}\label{sec:validity}\label{sec:multiband}

For a differentiable encoder, $\bG(\bp)$ can also be computed from its Jacobian (Equations~\ref{eq:jacobian} and~\ref{eq:quadratic-form}) by automatic differentiation, and we use such Jacobians as a reference to assess the fit in Section~\ref{sec:jacobian}.
Compared with the Jacobian, the tensor fit from finite-radius probes offers three advantages.
First, the fit requires only the output of the encoder, so it applies to models whose derivatives are not readily available. 
For example, the released CSP code computes its input features outside PyTorch, so automatic differentiation does not reach the input coordinates, and computing its Jacobians required reimplementing that step.
More generally, the fit requires no access to the implementation of an encoder, so it applies to encoders available only through their outputs, such as those served through an application programming interface or distributed as precomputed embeddings, as well as to encoders that are not differentiable.
Second, the Jacobian describes changes within an infinitesimal neighborhood, whereas the probe radius can be chosen to match the scale of interest. 
For ReLU-based networks, e.g., the Jacobian changes abruptly across activation-region boundaries, so its value at one location need not represent the response even over a short distance (supplementary Section~\ref{supp:relu}).
Third, the non-ellipticity $\nu$ (Equation~\ref{eq:nonellipticity}) quantifies the departure of directional responses from the estimated limiting stretch (the gap between the blue and red curves in Figure~\ref{fig:indicatrix}), whereas a Jacobian defines the tensor exactly and cannot reveal such departures.

The fitted tensor approximates $\bG(\bp)$, and its accuracy depends on the smoothness of the encoder within the probe disc.
For smooth encoders, $\widehat{\bG}_r$ converges to $\bG(\bp)$ as $r\to0$ (supplementary Section~\ref{supp:bound}).
This convergence, however, does not hold everywhere: when the probe disc crosses a non-smooth feature, such as an activation boundary of a ReLU network, the error need not shrink with the radius, and at very small radii floating-point rounding can dominate.
Coordinate discontinuities at the antimeridian or poles are another case where the encoder output can jump for minimal changes in input coordinates, so probe discs that cross such a discontinuity are excluded for the affected encoders (Section~\ref{sec:boundaries}; supplementary Section~\ref{supp:sampling}).

The choice of probe radius is therefore a trade-off between two sources of error. 
With a larger radius, the fitted tensor can be biased by the curvature of the feature response within the probe disc and, for ReLU-type encoders, by activation-region boundaries; with a smaller radius, the rounding error in the feature difference can be amplified as it is divided by $r$ in the finite stretch (Equation~\ref{eq:finite-stretch}; supplementary Section~\ref{supp:bound}).
Each family's primary radius (1\,km for GeoCLIP and SatCLIP, 0.3\,km for SINR, and 0.1\,km for CSP and Climplicit; Table~\ref{tab:enc}) was fixed before the assessments, within the range over which feature differences grew approximately in proportion to distance at test sites, and the same radius was used at every site.
To confirm that these radii lie in this range, we evaluated the curve of finite stretch over radii from 1\,m to 30\,km at four sites per checkpoint (supplementary Figure~\ref{fig:radius-screening}). 
The curves are flat around the primary radius except for GeoCLIP, whose stretch at 1\,km is 1--9\% below its small-radius value, so its results are also checked at 0.3\,km (Section~\ref{sec:jacobian}).

\section{Validation}\label{sec:validation}

Following Section~\ref{sec:validity}, we assess the estimator against two references: positional encodings whose geometry is known in closed form to test the estimator without a learned network (Section~\ref{sec:closed-form}), and model Jacobians of the trained encoders to test their response to infinitesimal displacements (Section~\ref{sec:jacobian}). Section~\ref{sec:boundaries} then identifies the impact of coordinate boundaries where the smoothness assumption cannot hold.

\subsection{Closed-form geographic references}\label{sec:closed-form}

Table~\ref{tab:refs} gives the infinitesimal anisotropy of the input encodings used by the measured families, and these expressions serve as input references in Section~\ref{sec:protocol}.
Except for Equal Earth, whose curved meridians make its anisotropy vary with both latitude and longitude, the anisotropy and principal bearing of the other input encodings depend only on latitude, to which we refer as latitude-only references.
To test the estimator itself, we apply it to these positional encodings directly (i.e., without a learned network) on the fixed $2.5^\circ\times5^\circ$ grid (supplementary Figure~\ref{fig:validation}). 
Every reference is probed at 1\,km, which lies within the range over which each input encoding changes linearly with distance.
The fitted anisotropy differs from the analytic value by at most 0.0012\% within $60^\circ$ latitude and by at most 0.61\% through $85^\circ$.
The largest difference occurs for the longitude-latitude sinusoids at $|\varphi|=85^\circ$, where $A\approx11.5$ and the east-west distance on the finest band shrinks to about 60\,km, consistent with the lower reliability of strongly anisotropic sites (Section~\ref{sec:finite}).
The SatCLIP input at $L=40$ is not included in these bounds.
Near the poles, its harmonics are computed in double precision from expanded polynomials whose large terms cancel, and the resulting numerical errors, mostly in the $\pi$-scaled zonal functions, make the input itself depart from its analytic form by up to 11\% in anisotropy (supplementary Section~\ref{supp:satclip}).

\begin{table}[htbp]
\centering\small
\caption{Infinitesimal anisotropy $A$ (greatest over least directional stretch) and major axis of the input encodings used as references (derivations in supplementary Section~\ref{app:closed}).}
\label{tab:refs}
\begin{tabularx}{\textwidth}{@{}>{\raggedright\arraybackslash}X>{\raggedright\arraybackslash}p{3.2cm}>{\raggedright\arraybackslash}p{2.9cm}>{\raggedright\arraybackslash}p{2.7cm}@{}}
\toprule
Positional encoding & Anisotropy ($A$) & Major axis & Used by\\\midrule
Longitude-latitude sinusoids with equal frequencies (Space2Vec grid) & $\sec\varphi$ & east-west & CSP\\[4pt]
Longitude and latitude scaled 2:1 & $\max(2\cos\varphi,\,\frac12\sec\varphi)$ & north-south for $|\varphi|<60^\circ$, east-west above & SINR, Climplicit\\[4pt]
Equal Earth coordinates, equally scaled & same as the Equal Earth projection & generally oblique; varies with longitude & GeoCLIP\\[4pt]
Spherical harmonics ($l<L$), $m=0$ components scaled by $c$ & $\sqrt{\max(q_{\rm SH},\,1/q_{\rm SH})}$ & north-south if $c>1$ & SatCLIP: released ($c=\pi$), input-rescaled ($c=1$)\\
\bottomrule
\end{tabularx}
\par\smallskip\parbox{\textwidth}{\footnotesize
$\varphi$: latitude. For each encoding, the north and east stretches are the norms of its derivatives with respect to latitude and longitude (in radians), divided by $R$ and $R\cos\varphi$ on a sphere of radius $R$; except for Equal Earth, these derivatives are orthogonal and $A$ is the ratio of the two stretches (e.g., $1/(R\cos\varphi)$ against $1/R$ for the Space2Vec grid, giving $\sec\varphi$). For Equal Earth, $A$ is that of the projection's Tissot indicatrix, computed from its analytic partial derivatives; GeoCLIP's random Fourier features add a median anisotropy of about 1.08 (Section~\ref{sec:geoclip}). $q_{\rm SH}$: northward divided by eastward squared stretch (supplementary Section~\ref{app:closed}); $c=1$ gives orthonormal harmonics, for which $A=1$. The SatCLIP implementation normalizes the $m=0$ components $\pi$ times larger than orthonormal harmonics, so $c=\pi$; this scaling can be absorbed into the first-layer weights and does not change the released outputs (supplementary Section~\ref{supp:satclip}).}
\end{table}

\subsection{Comparison with Jacobians}\label{sec:jacobian}

For differentiable encoders, the Jacobian can be computed by automatic differentiation, which can be used as a reference to check the accuracy of the fitted tensor.
In practice, the encoder, including any normalization of its outputs, is written as a function of latitude and longitude in radians, and an automatic-differentiation routine of the deep-learning framework (e.g., \texttt{torch.autograd.functional.jacobian} in PyTorch \citep{paszke2019pytorch}) returns the partial derivatives of all $d$ output components in one call; dividing the two resulting columns by $R$ and $R\cos\varphi$ gives $\bJ_{\bp}$ (Equation~\ref{eq:jacobian}).
Specifically, from the Jacobian $\bJ_{\bp}$ (Equation~\ref{eq:jacobian}), we calculate $\bG_J=\bJ_{\bp}^\top \bJ_{\bp}$ and its anisotropy $A_J$, and compare $A_J$ with the fitted anisotropy $A_r$ (Equation~\ref{eq:fitted-tensor}) of the same encoder output at its primary radius (Section~\ref{sec:validity}).
The comparison uses 2,340 sites through $80^\circ$ latitude for seven checkpoints, namely SatCLIP ViT-L10 and ViT-L40, Climplicit, GeoCLIP and the three SINR checkpoints, of which SINR cap-10 and distilled enter only the analysis of activation-region crossings below.
Agreement is close for SatCLIP and Climplicit, for which the 95th percentile of the relative approximation error $|A_r/A_J-1|$, weighted by site area (Section~\ref{sec:protocol}), is at most 0.24\%.
Errors are larger for SINR and GeoCLIP, with 95th percentiles of 1.15\% and 16.7\%, and rise to 16.1\% and 52.0\% at the most anisotropic sites (i.e., the upper decile of $A_J$). Supplementary Section~\ref{supp:jacobian} reports the full error distributions.
The non-ellipticity $\nu$ is related to these errors: its area-weighted median on the fixed grid ranges from $4.9\times10^{-5}$ (distilled SINR) to $6.4\times10^{-2}$ (GeoCLIP), and across the 2,340 sites, larger $\nu$ goes with larger anisotropy error for each of the five checkpoints compared above (Spearman rank correlation from 0.13 for GeoCLIP to 0.75 for SINR cap-1000), although a small $\nu$ does not guarantee a small error (Section~\ref{sec:finite}).

Both SINR and GeoCLIP use piecewise-linear ReLU activations in the network architecture.
When probe points fall in different activation regions from the center, the encoder outputs can change abruptly across the boundary, which violates the smoothness condition and may lead to larger errors.
For the three SINR checkpoints, only 6--12\% of discs contain probe points in a different activation region from the center, yet these discs account for every error above 0.1\%, whereas elsewhere the 95th-percentile error is at most $2\times10^{-5}$\%.
In GeoCLIP, in contrast, every disc contains crossings, and the error grows with the number of units that switch within the disc (supplementary Section~\ref{supp:relu}).
Reducing GeoCLIP's radius from 1 to 0.3\,km lowers the median number of switching units from 354 to 108 and, within $60^\circ$ latitude, the median anisotropy error from 3.90\% to 1.96\% and the 95th percentile from 11.92\% to 6.11\%.

For CSP, its released code builds the input features outside PyTorch and runs the network in single precision, which prevents automatic differentiation from reaching the input coordinates and makes the Jacobian sensitive to rounding error.
We re-implement its positional encoding with differentiable PyTorch operations, verify agreement with the released input features, and evaluate the same weights in double precision to compute Jacobians at 300 sites within $60^\circ$.
The fitted estimates at the primary 0.1\,km radius, from the released single-precision code, differ from these Jacobians by a median of 0.02\% (95th percentile 1.0\%) for CSP-iNat and 0.005\% (95th percentile 0.02\%) for CSP-fMoW (supplementary Section~\ref{supp:csp}).

\subsection{Coordinate boundaries}\label{sec:boundaries}

Coinciding or nearly coinciding locations should have nearly identical encoder output, even when their coordinates differ widely.
We test this property at the antimeridian, with pairs of points $10^{-6}$ degrees on either side of it at latitudes $0$, $45$ and $75^\circ$, and at the North Pole, described at longitudes $0^\circ$ and $-180^\circ$ (supplementary Table~\ref{tab:boundary}). 
Since outputs are unit-normalized, the distance within a pair ranges from 0 to 2.
Across the antimeridian, this distance is less than $10^{-6}$ for SINR and SatCLIP but 1.04--1.30 for GeoCLIP, 0.79--1.22 for Climplicit and 0.03--0.24 for CSP. At the North Pole, it is zero for SatCLIP and 0.05--1.15 for the other families.
A probe disc that spans the antimeridian would measure this jump rather than local geometry (Section~\ref{sec:validity}), so such discs are excluded for GeoCLIP, CSP and Climplicit.
The pole inconsistency does not affect the measurements, whose centers lie within $85^\circ$ latitude.

These results show that the estimator recovers closed-form geometry to within 0.0012\% within $60^\circ$ and 0.61\% through $85^\circ$, and it agrees with model Jacobians to within 0.24\% (95th percentile) for the smooth SatCLIP and Climplicit networks and to a median of 0.02\% or less for CSP evaluated in double precision. Larger departures arise where probe discs cross activation boundaries of ReLU networks, chiefly for GeoCLIP, and where coordinate discontinuities make probes measure a jump rather than local geometry, as for GeoCLIP, CSP and Climplicit across the antimeridian; the former limits site-level accuracy for GeoCLIP and SINR, and the latter discs are excluded from the analyses that follow.

\section{Local geometry of released encoders}\label{sec:encoders}

This section applies the estimator of Section~\ref{sec:theory}, validated in Section~\ref{sec:validation}, to the thirteen released checkpoints of Table~\ref{tab:encoders}. After describing the comparison protocol (Section~\ref{sec:protocol}), we assess the local geometry of the encoders and compare the output geometry of each family with its input reference (Sections~\ref{sec:geoclip}--\ref{sec:others}). 
We then test whether the fitted tensors predict which nearby locations an encoder treats as similar (Section~\ref{sec:neighborhoods}).

\subsection{Comparison protocol}\label{sec:protocol}\label{sec:sampling}

Anisotropy and bearing are invariant only to uniform transformations (Section~\ref{sec:setting}). 
Both unit normalization and the prediction head of a network are non-uniform transformations, so the measured distortion can depend on which feature vector is selected. For the three contrastive families of Table~\ref{tab:encoders} (GeoCLIP, SatCLIP and CSP), whose training compares cosine similarities, we use outputs scaled to unit length (denoted as U).
In contrast, SINR and Climplicit learn through prediction, and their unnormalized features feed a task-specific output layer; for these two families, we use the raw features (denoted as R) before that layer (i.e., the representation reused for other tasks), which for Climplicit concatenate the 256-dimensional outputs for months 3, 6, 9 and 12 into 1,024 components; the vector used for each checkpoint is stated in Table~\ref{tab:enc}.

We use a global grid with $2.5^\circ$ latitude and $5^\circ$ longitude spacing (4,968 sites, centers through $85^\circ$) for the assessment.
For the eight checkpoints whose anisotropy varies on scales finer than the grid spacing, as indicated by weak correlation between adjacent grid nodes (supplementary Section~\ref{supp:sampling}), we randomly displace the grid centers within their cells twice (9,936 sites) and repeat the measurement, so that the summaries are less sensitive to the placement of the grid, whereas the other five checkpoints use the fixed grid.
All summaries use the same estimator and are area-weighted, because the cells of a latitude-longitude grid shrink toward the poles in proportion to $\cos\varphi$.
To check that the comparison does not depend on the radius, we also re-evaluated every checkpoint at a common radius of 0.3\,km instead of its primary radius (Section~\ref{sec:validity}), which changes its median anisotropy by at most 0.03 ($A_{0.3\,km}$ in Table~\ref{tab:enc}), so the differences between families are not artifacts of the radius choice.
Stability under repeated halving of the radius and sensitivity to the number of bearings are reported in supplementary Section~\ref{supp:radius}.

We summarize the association between output and input anisotropy by $r_A$, the correlation between their log-anisotropies, $\log A$.
Agreement between the directions of greatest stretch is measured by $C_\theta=\langle\cos2(\theta_r-\theta_{\rm in})\rangle$, where $\theta_{\rm in}$ is the input-reference bearing and angle brackets denote a mean.
$C_\theta$ equals 1 for aligned axes, 0 for axes separated by $45^\circ$, and $-1$ for perpendicular axes.
The two statistics describe different properties, since directions can align even when anisotropy magnitudes vary differently across space. 
Both use sites within $60^\circ$ latitude, and their 95\% intervals are derived from geographic block resampling, in which latitude-longitude blocks of sites are drawn with replacement and the statistics recomputed 1,000 times (supplementary Section~\ref{supp:sampling}).
The input reference is the geometry expected in the output if the network added no directional distortion of its own (i.e., if it were locally conformal on the input surface), so $r_A$ and $C_\theta$ measure association of location encoders with this input.
Among the input references, only GeoCLIP's Equal Earth reference varies with longitude (Section~\ref{sec:closed-form}).
The other twelve vary only with latitude and, within $60^\circ$, have a constant cardinal direction of greatest stretch (north-south for SINR, Climplicit and SatCLIP; east-west for CSP), hence for them $C_\theta$ actually measures an overall preference for one cardinal axis across the globe.

\begin{table}[htbp]\centering\small
\caption{Geometry of released checkpoints and association with input references. Bracketed intervals are 95\% geographic block-resampling intervals, which describe sensitivity to the regions sampled.}
\label{tab:enc}
\begin{adjustbox}{max width=\textwidth}
\begin{tabular}{@{}lrcrrrr@{}}
\toprule
Encoder & Radius (km)&Vector&Median $A$&$A_{0.3\,km}$ & $r_A$ [95\% interval] & $C_\theta$ [95\% interval]\\\midrule
GeoCLIP&1&U&1.46&1.49&0.88 [0.79, 0.90]&0.88 [0.85, 0.92]\\
SatCLIP ViT-L10&1&U&1.57&1.57&0.10 [$-$0.01, 0.21]&0.49 [0.40, 0.58]\\
SatCLIP R18-L10&1&U&1.50&1.50&0.13 [0.02, 0.27]&0.49 [0.39, 0.59]\\
SatCLIP R50-L10&1&U&1.50&1.50&0.13 [0.03, 0.25]&0.55 [0.44, 0.65]\\
SatCLIP ViT-L40$^\dagger$&1&U&1.44&1.44&0.02 [$-$0.04, 0.08]&$-$0.01 [$-$0.07, 0.05]\\
SatCLIP R18-L40$^\dagger$&1&U&1.44&1.44&0.03 [$-$0.04, 0.08]&0.02 [$-$0.03, 0.09]\\
SatCLIP R50-L40$^\dagger$&1&U&1.43&1.42&0.04 [$-$0.04, 0.10]&0.02 [$-$0.05, 0.08]\\
SINR cap-1000$^\dagger$&0.3&R&2.89&2.91&0.09 [0.02, 0.17]&0.38 [0.31, 0.45]\\
SINR cap-10&0.3&R&2.99&2.99&$-$0.19 [$-$0.35, $-$0.01]&0.44 [0.31, 0.56]\\
SINR distilled$^\dagger$&0.3&R&4.88&4.82&0.06 [$-$0.02, 0.15]&0.25 [0.18, 0.32]\\
CSP-iNat$^\dagger$&0.1&U&2.44&2.50&$-$0.07 [$-$0.11, $-$0.02]&0.19 [0.15, 0.23]\\
CSP-fMoW$^\dagger$&0.1&U&1.81&1.80&0.23 [0.17, 0.28]&0.14 [0.08, 0.19]\\
Climplicit$^\dagger$&0.1&R&1.77&1.77&0.19 [0.13, 0.24]&0.56 [0.52, 0.60]\\
\bottomrule\end{tabular}
\end{adjustbox}
\par\smallskip\parbox{.97\textwidth}{\footnotesize U: unit-normalized output; R: raw features. Median $A$: fitted anisotropy $A_r$ at the primary radius over $|\varphi|\le85^\circ$, including flagged fallbacks; medians are descriptive and do not rank model quality. $A_{0.3\,km}$: median anisotropy re-evaluated at a common 0.3\,km radius on the fixed grid. $^\dagger$Summaries pooled over two randomized grids, which accounts for differences of up to 0.06 between the two median columns; other rows use the fixed grid. $r_A$ and $C_\theta$: $|\varphi|\le60^\circ$; $C_\theta$ uses only sites where both anisotropies exceed $1+10^{-6}$. Probe discs crossing the antimeridian are excluded for GeoCLIP, CSP and Climplicit. SatCLIP input references are the implemented ($\pi$-scaled) harmonic inputs, probed at 1\,km without unit normalization.}
\end{table}
\relax\space%

\subsection{GeoCLIP: association with Equal Earth}\label{sec:geoclip}

\begin{figure}[htbp]
\centering\includegraphics[width=\textwidth]{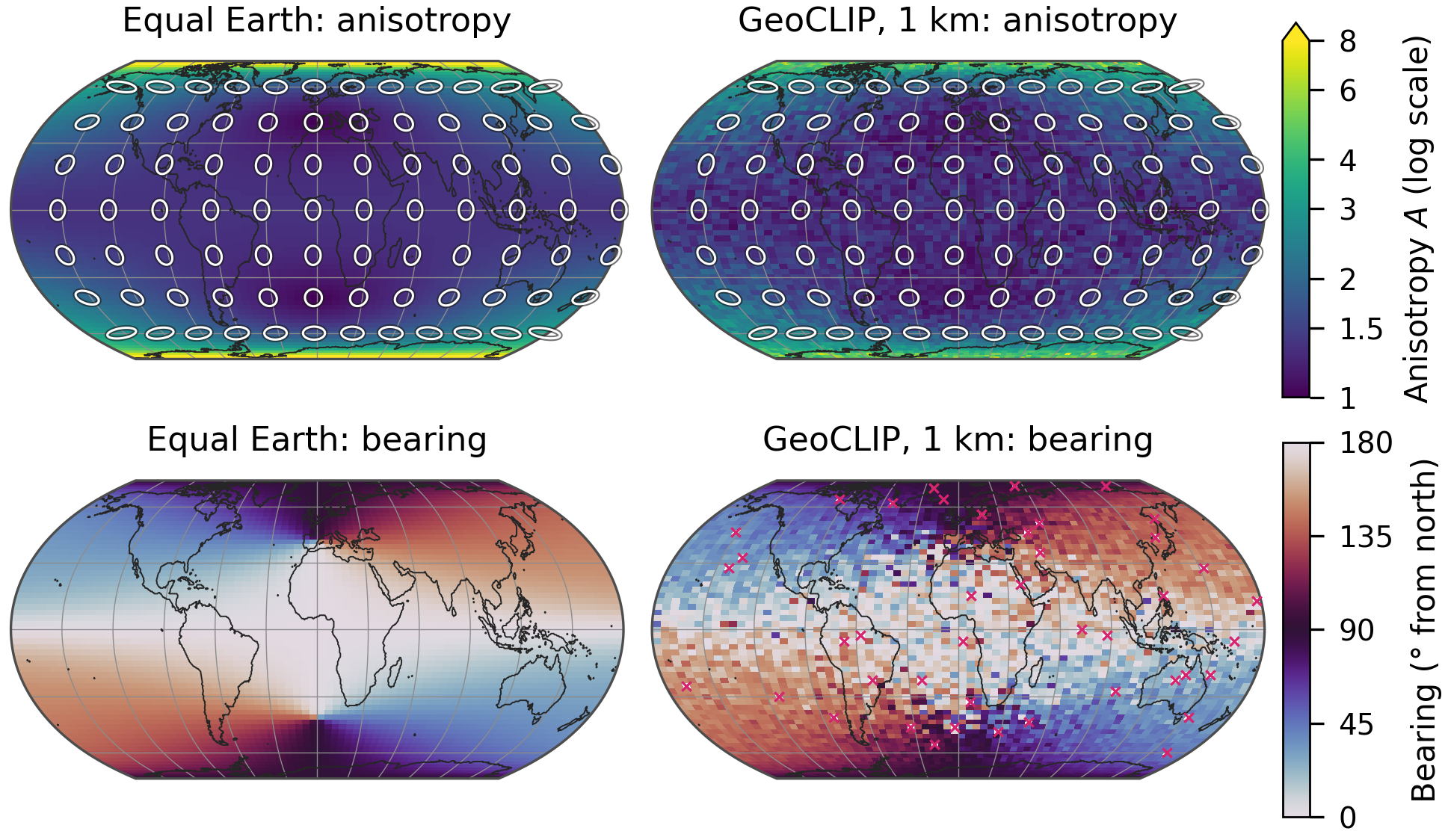}
\caption{Analytic Equal Earth reference (left) and GeoCLIP at 1\,km (right) on the fixed $2.5^\circ\times5^\circ$ grid, mapped in the Equal Earth projection. Top: anisotropy $A$ itself (not $\log A$) on a logarithmic color scale shared with Figures~\ref{fig:dense} and~\ref{fig:satclip-backbones}, saturating at $A=8$. White ellipses, every $30^\circ$ of longitude and $20^\circ$ of latitude within $60^\circ$, are drawn in the manner of Tissot's indicatrix (Figure~\ref{fig:schematic}a) at a common size in each site's north-east frame and show only the axis ratio $A$ and the principal bearing (along the long axis); for the Equal Earth reference, they have the shape of its Tissot indicatrices but follow the principal bearing on the ground rather than on the map. Bottom: principal bearing clockwise from north, modulo $180^\circ$. Pink crosses mark sampled-minimum fallbacks in GeoCLIP; their bearings are masked.}
\label{fig:geoclip}
\end{figure}

GeoCLIP first projects geographic coordinates using Equal Earth, and then combines three random Fourier feature branches with different frequency settings through learned networks.
Its input reference is therefore the Equal Earth stage, although the random frequency matrices add slight anisotropy of their own (a median of about 1.08 for 256 Gaussian frequencies; supplementary Section~\ref{supp:geoclip}).
Because Equal Earth is equal-area, its Tissot distortion is entirely distortion of shape, which is the quantity that $A$ and bearing measure.
At a probe radius of 1\,km, the output log-anisotropy correlates 0.88 with the analytic Equal Earth log-anisotropy, and $C_\theta$ is 0.88 (Table~\ref{tab:enc}). 
The tensor fitting error is unlikely to contribute to this association, since on the 1,764 Jacobian sites within $60^\circ$, the correlation is 0.869 when the tensors are computed from Jacobians against 0.874 for the fitted tensors (supplementary Section~\ref{supp:geoclip}).
Figure~\ref{fig:geoclip} shows similar geographic patterns, which visually confirm this association.

The association includes not only a common latitude dependence but also longitudinal structure.
Equal Earth's oblique meridians produce a longitude-dependent reference, and after subtracting each latitude row's mean log-anisotropy from both fields, the correlation remains 0.81 within $60^\circ$.
A possible reason is that Fourier-feature networks learn with a kernel that is approximately stationary and isotropic in the input coordinates  \citep{tancik2020fourier,jacot2018}, so a network whose sensitivity is nearly equal in each direction inherits Equal Earth's shape distortion on the sphere.

\subsection{SatCLIP: input and learned geometry}\label{sec:satclip}

Compared with GeoCLIP, SatCLIP shows weaker association with its input reference.
The three $L=10$ checkpoints show a tendency toward alignment with the reference's north-south axis ($C_\theta$ of 0.49--0.55), while the $L=40$ checkpoints show little net alignment with that axis ($C_\theta$ from $-0.01$ to 0.02; Table~\ref{tab:enc}).
Anisotropy correlations are small for both degree limits (0.10--0.13 for $L=10$ and 0.02--0.04 for $L=40$). 
For $L=40$, the anisotropy of the input reference varies only from 1.09 to 1.18 within $60^\circ$, which leaves little variation for $r_A$ to detect.
At each degree limit, checkpoints trained with different image encoders produce similar anisotropy maps (supplementary Figure~\ref{fig:satclip-backbones}), and the oscillations in the $L=40$ fields (Figure~\ref{fig:dense}) match the harmonic patterns reported by \citet{rao2026intrinsic}.

In the released checkpoints, the zonal ($m=0$) components of the harmonic input are scaled by $\pi$ relative to orthonormal harmonics (supplementary Section~\ref{supp:satclip}).
This scaling factor could be absorbed into the first-layer weights without changing the released outputs, but it changes the input reference used for $r_A$ and $C_\theta$.
To test how much of this anisotropy comes from the $\pi$ scaling, we rescaled the zonal ($m=0$) input components to the orthonormal harmonics while holding the weights of all six networks fixed. 
At 35 selected sites, the rescaling lowers the median output anisotropy of the three $L=10$ checkpoints (ViT from 1.68 to 1.54, R18 from 1.58 to 1.42 and R50 from 1.51 to 1.37).
In contrast, the $L=40$ checkpoints show no consistent change (supplementary Table~\ref{tab:satclipcorrection}).
These results are based on the implementation used by the released checkpoints at the time of writing, and a corrected implementation may change the input reference.

\subsection{Other families and sampling sensitivity}\label{sec:others}

\begin{figure}[htbp]
\centering\includegraphics[width=\textwidth]{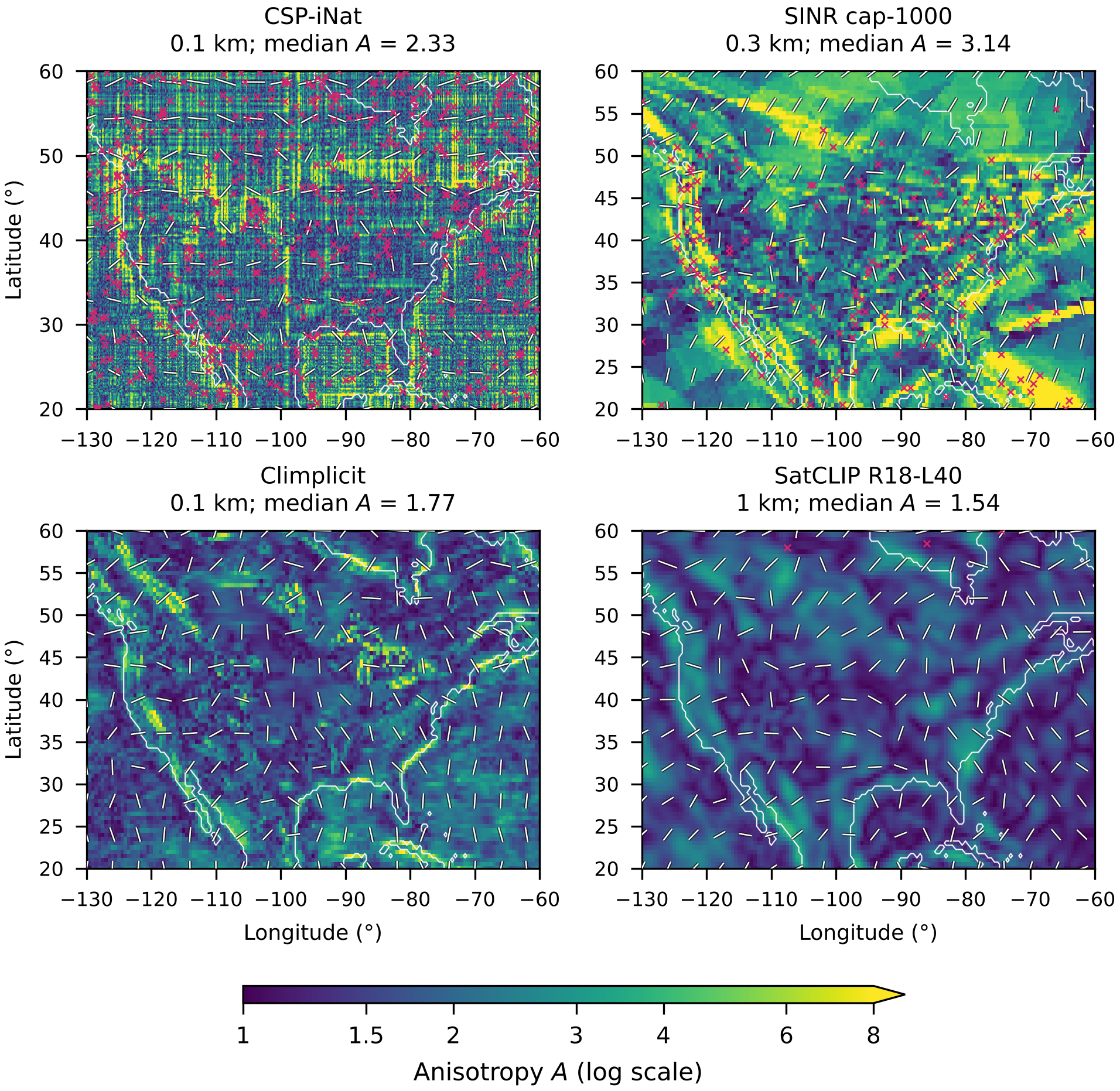}
\caption{Local anisotropy and principal directions over North America ($20$--$60^\circ$N, $130$--$60^\circ$W) for four released checkpoints. CSP is sampled at $0.1^\circ$, and SINR, Climplicit and SatCLIP R18-L40 at $0.5^\circ$. Color shows $A$ itself (not $\log A$) on the logarithmic color scale shared with Figures~\ref{fig:geoclip} and~\ref{fig:satclip-backbones}; values above 8 share the top color. Line segments show the fitted principal bearing; pink crosses mark sampled-minimum fallbacks, where bearing segments are omitted. Each panel uses an aspect ratio of $1/\cos40^\circ$, so local north-south and east-west distances at the central latitude are drawn to the same scale.}
\label{fig:dense}
\end{figure}

As with SatCLIP, the SINR, CSP and Climplicit checkpoints show weak anisotropy associations with their input references and spatial patterns beyond the latitude dependence of those references.
As shown in the dense regional maps (Figure~\ref{fig:dense}), CSP exhibits fine rectilinear patterns, SINR straight streaks, and Climplicit spatially varying anisotropy and orientation.
The CSP pattern corresponds to the grid pattern that \citet{rao2026intrinsic} observed in local intrinsic dimension, which is represented here as anisotropy and orientation.
The SINR streaks may follow boundaries between activation regions of its ReLU network, where its finite-probe errors increase (Section~\ref{sec:jacobian}). %

For these checkpoints, $|r_A|$ is at most 0.23, and two of them, SINR cap-10 and CSP-iNat, have negative $r_A$, which indicates that their output anisotropy tends to be lower where the input reference is higher.
Because latitude explains at most 17\% of the variance of output log-anisotropy for these checkpoints (supplementary Figure~\ref{fig:latitude}), $r_A$ can remain small even where latitude profiles agree.
These weak associations are also sensitive to grid placement in finely structured fields, particularly for CSP (supplementary Section~\ref{supp:sampling}), and to the inclusion of ocean sites, since restricting the sample to land leaves GeoCLIP's association unchanged but changes several latitude-only associations (supplementary Table~\ref{tab:land}).

\subsection{Local distortion and geographic neighborhoods}\label{sec:neighborhoods}

Finally, we examine the departure of neighborhoods defined by embedding similarity from geographic neighborhoods, and whether the fitted tensor, which defines the reach ellipse of Section~\ref{sec:setting}, accounts for it.
The comparison covers twenty query locations between $0^\circ$ and $70^\circ$N (latitudes $0^\circ$, $30^\circ$, $45^\circ$, $60^\circ$ and $70^\circ$N at longitudes $0^\circ$, $60^\circ$, $120^\circ$ and $170^\circ$E) and fifteen model variants, including the thirteen released checkpoints and the two SatCLIP ViT variants with rescaled zonal inputs (Section~\ref{sec:satclip}), giving a total of 300 comparisons, of which 299 are valid after excluding one fallback for distilled SINR.
At each query, we draw 400 candidate locations uniformly by area within a 1\,km disc and select forty neighbors from them in three ways, namely geographic neighbors that minimize geographic distance, embedding neighbors that minimize the actual feature distance between the measured vectors of Section~\ref{sec:protocol}, and tensor neighbors that minimize the predicted feature distance $\sqrt{\bdelta^\top\widehat{\bG}_r\bdelta}$, with $\widehat{\bG}_r$ the fitted tensor of Equation~\ref{eq:fitted-tensor} and $\bdelta$ the geographic displacement from the query.

\begin{figure}[htbp]\centering
\includegraphics[width=\textwidth]{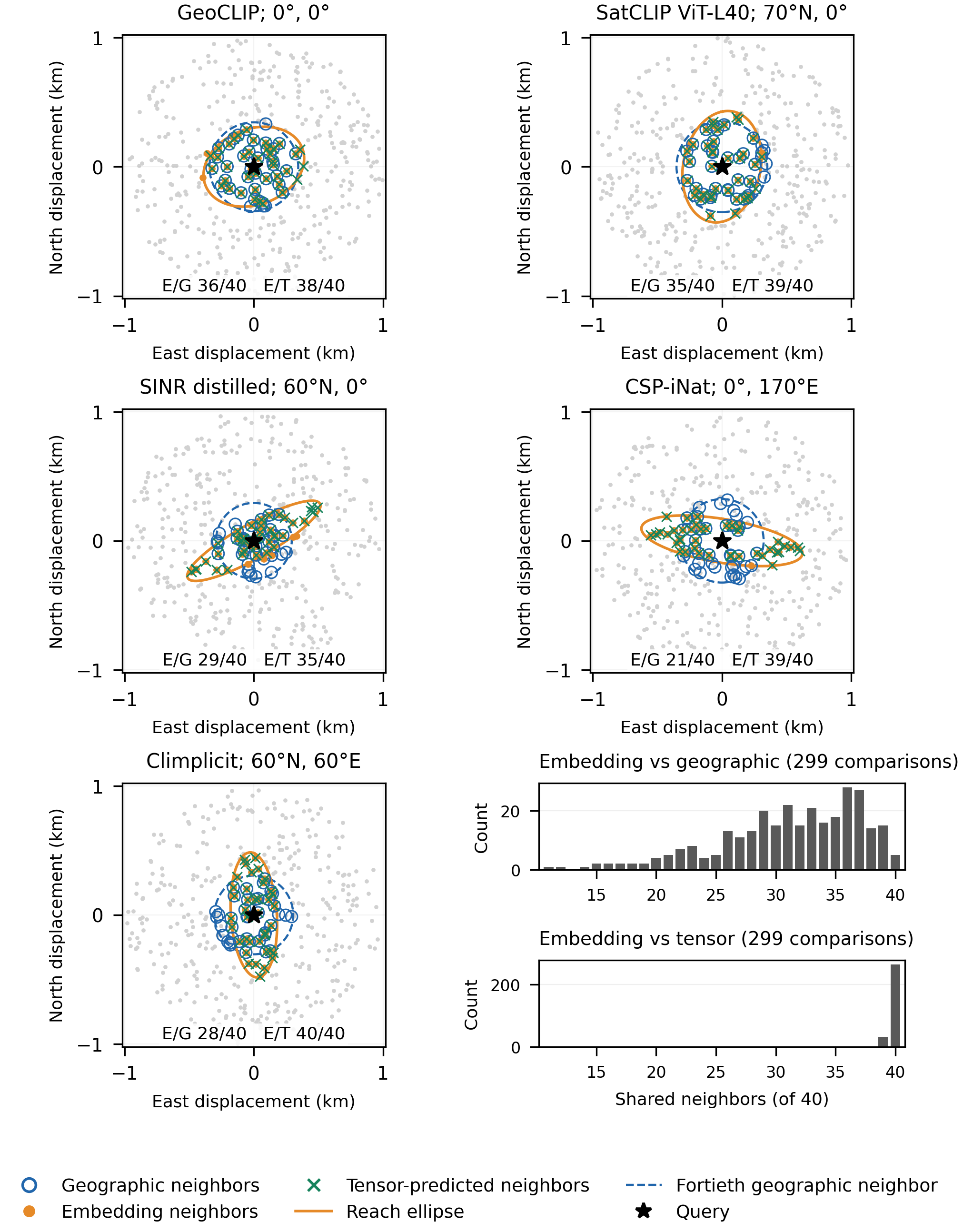}
\caption{Embedding neighborhoods and their tensor approximation. Each map selects forty neighbors among 400 candidates within 1\,km of the query. The dashed circle passes through the fortieth geographic neighbor; the reach ellipse is the contour of tensor-predicted feature distance at the actual fortieth-neighbor threshold. E/G and E/T give embedding-geographic and embedding-tensor overlap (T/G in supplementary Table~\ref{tab:neighbourhoods}). The maps include the two comparisons with the lowest E/T overlap. Histograms count shared neighbors over all 299 valid comparisons; forty means identical selections.}
\label{fig:neighborhoods}
\end{figure}

For every checkpoint, we fit the tensor with Equations~\ref{eq:profile-fit}--\ref{eq:fitted-tensor} from separate directional probes at a common radius of 0.3\,km (rather than the family-specific radii in Table~\ref{tab:enc}).
Overlap is the fraction of the forty neighbors that two selections share (E/G for embedding and geographic neighbors, E/T for embedding and tensor neighbors). 
In the 1\,km disc, the forty embedding neighbors extend to a median of 0.40\,km, close to the fitting radius.
Figure~\ref{fig:neighborhoods} maps the results for one checkpoint per family.
In these maps, the reach ellipse (Section~\ref{sec:setting}) is drawn from the fitted tensor and its size from the actual fortieth-neighbor distance, and it therefore need not enclose exactly forty candidates.

Even within 1\,km, embedding neighbors depart substantially from geographic neighbors: median E/G overlap ranges from 57.5\% to 87.5\% across the thirteen released checkpoints and is 85.0\% and 90.0\% for the two rescaled SatCLIP variants, and geographic distance selects the identical forty neighbors in only 5 of 299 comparisons (further alternatives in supplementary Table~\ref{tab:neighbournulls}).
The fitted tensor largely explains this departure, with a median E/T overlap of 97.5\% for GeoCLIP and 100\% for all other variants, and identical selections in 265 of 299 (88.6\%) valid comparisons.
Since the neighbors at this scale lie close to the fitting radius, such agreement is expected where the linear approximation holds, and it mainly confirms that the measured tensor captures the local response of each encoder.
The agreement also persists beyond the fitting scale; in 6\,km discs, where the forty embedding neighbors extend to a median of 2.4\,km (eight times the fitting radius), the tensor still selects the identical set in 225 of 299 comparisons (supplementary Table~\ref{tab:neighbourextrap}).
It should be noted that the comparison is conducted on a limited number of query sites rather than a global assessment, and the sensitivity to candidate radius and neighbor count is reported in supplementary Section~\ref{supp:neighborhoods}.

In summary, the thirteen checkpoints differ widely in median anisotropy for the feature vectors used (Table~\ref{tab:enc}), from 1.43 for SatCLIP R50-L40 (unit-normalized) to 4.88 for distilled SINR (raw features), and only GeoCLIP closely follows its input reference from site to site ($r_A=0.88$, $C_\theta=0.88$).
The other checkpoints show weak anisotropy associations with input ($|r_A|\le0.23$) but varying bearing alignment, from no clear alignment for SatCLIP $L=40$ ($C_\theta$ between $-0.01$ and 0.02) to $C_\theta=0.56$ for Climplicit, and rescaling SatCLIP's zonal harmonics lowers the anisotropy of its $L=10$ checkpoints at 35 selected sites.
Despite these differences, embedding neighborhoods depart from geographic ones for every family, and the fitted tensors largely account for this departure, selecting exactly the same forty neighbors as the embedding in 265 of 299 comparisons within 1\,km (225 within 6\,km), against 5 for geographic distance.

\section{Discussion}\label{sec:discussion}

\subsection{Geographic information in representation geometry}\label{sec:geoinfo}

The results in Section~\ref{sec:neighborhoods} show that the locations closest in an embedding space need not be the closest in geographic distance.
As the fitted tensor recovers the embedding neighbors in most comparisons, the reach ellipse can be taken as the geographic reach of embedding similarity at a common predicted feature distance, giving embedding neighborhoods a geographic interpretation.
This interpretation extends Tissot's indicatrix from map projections to high-dimensional embedding spaces.
From the perspective of spatial statistics, the reach ellipse also connects to deformation models of nonstationary covariance \citep{sampson1992,paciorek2006,risser2015}, in which an isotropic covariance in a deformed space becomes locally anisotropic in geographic space. 
Location encoders can therefore be interpreted as learned deformations for spatial prediction.
Furthermore, the association of GeoCLIP with Equal Earth, and the departure of SatCLIP, SINR, CSP and Climplicit from their input references, indicate that the input coordinate construction does not necessarily determine the learned geometry of location encoders.

\subsection{Implications for the assessment and development of location encoders}\label{sec:practice}\label{sec:development}

Our results have several implications for the assessment of location encoders.
First, predictive performance alone (e.g., error statistics), the usual basis for evaluating location encoders (Section~\ref{sec:evaluation}), does not reveal how an encoder represents nearby locations. 
As Section~\ref{sec:neighborhoods} shows, embedding neighbors and geographic neighbors share a median of only 57.5--87.5\% of their members within the tested 1\,km discs.
For downstream applications that rely on embedding similarity, such as retrieval and neighbor-based prediction, reporting local distortion alongside predictive performance can help interpret the results.
For EO applications in particular, labels can be transferred from the nearest neighbors in a feature or embedding space \citep{chirici2016knn,brown2025alphaearth}, whereas spatial cross-validation separates training and test data by geographic distance \citep{roberts2017cross}, and the two notions of proximity can disagree (Section~\ref{sec:neighborhoods}).
Second, the local distortion of an encoder depends on the feature vector (e.g., normalized or unnormalized) and the related distance or similarity measure, such as Euclidean distance or cosine similarity (Section~\ref{sec:protocol}), and an assessment should therefore state the type of vectors and measures actually used.
Third, assessments should report anisotropy and principal bearing with the probe radius, and describe the smooth characteristics, as finite-distance estimates for non-smooth encoders (e.g., piece-wise linear encoders such as ReLU networks) can depart substantially from their Jacobians (Section~\ref{sec:jacobian}).

In addition, the results have implications for the development of location encoders.
First, GeoCLIP's learned geometry is strongly associated with its Equal Earth reference (Section~\ref{sec:geoclip}), which motivates controlled comparisons of the coordinate systems used for positional encoding. 
The present association does not establish how the projection, architecture and training produced that geometry.
Second, encoder implementations should be checked for the numerical accuracy of input features and for the treatment of the antimeridian and poles.
For example, high-degree spherical harmonics evaluated from expanded polynomials are computed with large errors near the poles, as double precision floating point cannot resolve the cancellation of their large terms (supplementary Section~\ref{supp:satclip}).
In addition, several released encoders give different outputs on the two sides of the antimeridian or at the North Pole (Section~\ref{sec:boundaries}), which a simple continuity check can detect.
Third, piecewise-linear activations make the local geometry well defined only within activation regions, and the finite-probe estimates of ReLU networks depart more from their Jacobians than those of smooth networks (Section~\ref{sec:jacobian}).
It should be noted, however, that isotropy is not necessarily a design or quality target, since, as in cartography, no single map projection suits every purpose; conformal projections preserve local shape (i.e., $A=1$ everywhere) at the cost of distorting area, while equal-area projections such as Equal Earth accept shape distortion in order to preserve area \citep{snyder1987}.
Similarly, an elongated neighborhood in a location encoder may reflect environmental variation useful for prediction, and whether reshaping the measured geometry improves a model requires task-specific evaluation.

\subsection{Limitations and future work}\label{sec:limits}

This study has several limitations.
First, the measurements describe local geometry at a specified probe radius, and assume that the encoder is smooth over the probe neighborhood.
This assumption may fail particularly when probe discs cross activation boundaries of ReLU networks or where single-precision rounding dominates (Section~\ref{sec:jacobian}; supplementary Sections~\ref{supp:bound} and~\ref{supp:relu}), and the independent Jacobian checks are only available for nine of the thirteen checkpoints.
Second, we did not attempt to attribute the distortions to the architecture or the training process of the location encoders; understanding the attribution needs controlled experiments, for example retraining the same architecture on different data.
Third, the results describe thirteen released checkpoints from five families at one primary radius each, and other encoders and probe radii may behave differently. 
Finally, the neighborhood comparison covers areas up to 6\,km of twenty selected queries between $0^\circ$ and $70^\circ$N, and whether the agreement holds over larger neighborhoods, and whether the measured distortion affects downstream tasks, remain to be tested.
In future work, we will examine these consequences through downstream case studies with encoders trained on Earth-observation and other imagery (e.g., SatCLIP, CSP and GeoCLIP), such as nearest-neighbor prediction of land cover or species occurrence and image-to-location retrieval, and test whether neighborhoods shaped by the reach ellipse, rather than isotropic geographic buffers, improve such predictions.
Since the estimator requires only encoder outputs, another topic that warrants further investigation is the assessment of precomputed embedding products beyond location encoders, such as global satellite embeddings \citep{brown2025alphaearth}, whose models are not directly accessible.

\section{Conclusion}\label{sec:conclusion}

In this paper, we adapt Tissot's indicatrix to characterize local distortion in learned location embeddings.
Specifically, we first fit a tensor to the encoder's response to equal-distance geodesic displacements and validate it against analytic references and model Jacobians.
We then characterize the local distortions of thirteen released checkpoints from five encoder families relative to the input reference implied by each encoder's coordinate construction.
GeoCLIP is strongly associated with its Equal Earth reference ($r_A=0.88$, $C_\theta=0.88$), whereas the other families show only weak anisotropy associations ($|r_A|\le0.23$) and varying degrees of bearing alignment.
The selected input references therefore do not generally predict the output geometry, and the directional response needs to be measured for the specified learned representation.
The measurements also reveal implementation issues at the time of writing, namely a $\pi$ scaling of zonal harmonics in SatCLIP, which makes its input anisotropic, numerical errors in its high-degree harmonics near the poles, and output discontinuities at the antimeridian or the North Pole in several released encoders.
We further examine the consequence for geographic neighborhoods by comparing the neighbors selected by geographic distance, embedding distance and tensor-predicted distance, which shows that embedding neighbors depart substantially from geographic neighbors (a median overlap of 57.5--87.5\% within 1\,km), that the fitted tensor largely accounts for this departure (identical neighbor sets in 265 of 299 comparisons within 1\,km and 225 within 6\,km), and that the reach ellipse describes the directional reach of embedding similarity.

These findings provide a geographic basis for assessing and developing location encoders, and the eTissot package makes the measurement available for a specified encoder, location and probe radius.
The consequences of the measured distortion for downstream tasks, and the respective contributions of architecture and training to the measured geometry, are topics that warrant further investigation.

\section*{Disclosure statement}

The authors declare that they have no known competing financial interests or personal relationships that could have appeared to influence the work reported in this paper.

\section*{Funding}

Guofeng Cao was supported by the U.S. National Science Foundation (NSF, Award No.~2026331) and the National Aeronautics and Space Administration (NASA, Award No.~80NSSC26K1405). Gengchen Mai was supported by the NSF (Award No.~2521631) and the Good Systems Smart Cities project at the University of Texas at Austin.

\section*{Data and codes availability statement}

The accompanying eTissot package (\texttt{etissot}) is openly available at \url{https://github.com/surfcao/etissot} (version 0.3.3).
It contains the code to conduct the proposed finite-probe measurement for a user-supplied encoder at a specified location and probe radius, and to reproduce the reported results and figures in the manuscript.

\section*{AI usage disclosure}

We used AI tools (Claude Opus 5.5, Anthropic) for coding assistance in developing the eTissot package and for language editing and revision of the manuscript. 
All AI-assisted code and text were reviewed, tested and revised by the authors, who take full responsibility for the content of the article.

\bibliographystyle{plainnat}
{\interlinepenalty=10000\widowpenalty=10000\clubpenalty=10000\bibliography{references}}

\clearpage
\phantomsection
\section*{Supplementary material}
\addcontentsline{toc}{section}{Supplementary material}
\setcounter{section}{0}
\setcounter{subsection}{0}
\setcounter{figure}{0}
\setcounter{table}{0}
\setcounter{equation}{0}
\setcounter{proposition}{0}
\renewcommand{\thesection}{S\arabic{section}}
\renewcommand{\thefigure}{S\arabic{figure}}
\renewcommand{\thetable}{S\arabic{table}}
\renewcommand{\theequation}{S\arabic{equation}}
\renewcommand{\theHsection}{supp.\arabic{section}}
\renewcommand{\theHsubsection}{\theHsection.\arabic{subsection}}
\renewcommand{\theHfigure}{supp.\arabic{figure}}
\renewcommand{\theHtable}{supp.\arabic{table}}
\renewcommand{\theHequation}{supp.\arabic{equation}}
\def\theHproposition{supp.\arabic{proposition}}

\section{Local geometry, estimation and error control}\label{supp:geometry}

This section first derives the harmonic form of the directional profile used in main Section~\ref{sec:finite} and describes the fallback rule for the fitted tensor, then summarizes the sources of estimation error, and finally gives the closed-form input references used in main Section~\ref{sec:closed-form}.

\subsection{Harmonic form and tensor fit}\label{supp:harmonic}
\paragraph{Harmonic form and asymmetry} For a differentiable encoder $\mathbf{f}$, the local expansion $\mathbf{f}(\bp_r(\theta))-\mathbf{f}(\bp)=r\bJ_{\bp}\bu(\theta)+o(r)$, with $\bu(\theta)=(\cos\theta,\sin\theta)^\top$ and bearing $\theta$ measured clockwise from north, gives the limiting squared stretch (with $\bp$ omitted for brevity)
\[
s_0^2(\theta)=\|\bJ_{\bp}\bu(\theta)\|_2^2
=\bu(\theta)^\top \bG\bu(\theta)
=G_{NN}\cos^2\theta+2G_{NE}\sin\theta\cos\theta+G_{EE}\sin^2\theta.
\]
Substituting $\cos^2\theta=(1+\cos2\theta)/2$, $\sin^2\theta=(1-\cos2\theta)/2$ and $2\sin\theta\cos\theta=\sin2\theta$ yields main Equation~\ref{eq:limiting-stretch}, $s_0^2(\theta)=a_0+a_2\cos2\theta+b_2\sin2\theta$, with $a_0=(G_{NN}+G_{EE})/2$, $a_2=(G_{NN}-G_{EE})/2$ and $b_2=G_{NE}$. As this limiting profile has no first harmonic, the first-harmonic terms are added to main Equation~\ref{eq:profile-fit} to account for the asymmetry between opposite directions that arises from the finite probe radius.
This asymmetry comes from the variation of the encoder's derivatives within the probe disc: for a twice-differentiable encoder, the next term in the expansion of $s_r^2(\theta)$ is proportional to $r$ and changes sign between opposite directions, so the first harmonic vanishes as $r\to0$ (Section~\ref{supp:bound} gives the order of the remaining error).

The tensor is fitted by ordinary least squares with the columns $[1,\cos2\theta,\sin2\theta,\cos\theta,\sin\theta]$ on 32 equally spaced bearings.
Because the 32 bearings are equally spaced over the full circle and equally weighted, the first-harmonic columns are orthogonal to the constant and double-angle columns, so the tensor coefficients are the same whether or not the first harmonic is included in the fit.
It should be noted that the first harmonic still enters the non-ellipticity $\nu$ (main Equation~\ref{eq:nonellipticity}), which measures residuals from $\widehat{s_0^2}(\theta)$ alone.
On the fixed grid, it accounts for an area-weighted median of at least 68\% of the mean squared residual for every checkpoint (on equally spaced bearings, the residual splits exactly into the first-harmonic part, $(\widehat a_{1,r}^2+\widehat b_{1,r}^2)/2$, and the remainder), which indicates that most of the misfit measured by $\nu$ is asymmetry between opposite directions and does not affect the fitted tensor (main Equation~\ref{eq:fitted-tensor}).

\paragraph{Fallback for the minor value} From the fitted coefficients, the eigenvalues of $\widehat{\bG}_r$ are $\widehat\lambda_{1,r},\widehat\lambda_{2,r}=\widehat a_{0,r}\pm\sqrt{\widehat a_{2,r}^2+\widehat b_{2,r}^2}$.
As anisotropy divides by the minor value $\widehat\lambda_{2,r}$, a fitted minor value that is too small can produce extremely large anisotropy, and hence needs to be checked.
For an exact ellipse, the fitted minor value is smaller than the measured squared stretch at other bearings.
Specifically, the fitted squared stretch rises from $\widehat\lambda_{2,r}$ on the minor axis and reaches $\widehat\lambda_{2,r}\{1+(A_r^2-1)\sin^2\epsilon\}$ at angle $\epsilon$ from it. 
For the 32 equally spaced bearings, the closest bearing to the minor axis is at most $\pi/32$ away, so the smallest measured squared stretch should not exceed the fitted value at that bearing, which gives the lower bound:
\[
\widehat\lambda_{2,r}\{1+(A_r^2-1)\sin^2(\pi/32)\}.
\]
When the profile departs substantially from an ellipse, however, the least-squares fit can place the minor value further below the measurements than this bound allows, or even below zero.
The fitted minor value is therefore retained only when it is positive and the smallest measured squared stretch exceeds this bound by at most a relative tolerance of $10^{-6}$, which allows for rounding (fallback counts on the fixed grid are unchanged for tolerances from 0 to $10^{-6}$).
Otherwise, a flagged sampled-minimum fallback retains the major value and bearing, replaces $\widehat\lambda_{2,r}$ by $\max(\min_j y_j,\,10^{-12}\widehat\lambda_{1,r})$ and recomputes $A_r$, where the floor $10^{-12}\widehat\lambda_{1,r}$ keeps $A_r$ finite.
Because both thresholds scale with the squared stretches, the fallback is unaffected by a common rescaling of the outputs.
The fallback tends to underestimate anisotropy, considering the lower bound of the sampled squared stretch.
Fallbacks occur at fewer than 1\% of sites for every checkpoint and are included in main Table~\ref{tab:enc}; excluding them changes $r_A$ by at most 0.001.

\subsection{Finite-radius, rounding and activation-boundary error}\label{supp:bound}
\paragraph{Smooth encoders} Where the encoder changes smoothly within the probe disc, the fitted tensor $\widehat{\bG}_r$ approaches $\bG(\bp)$ as the radius shrinks, and two properties of the fit determine how errors in the measured profile propagate to the estimates.
First, on 32 equally spaced bearings, the fitted squared stretch $\widehat{s_0^2}(\theta)$ in any direction is a weighted sum of the measured squared stretches, with weights summing to one; because some weights are negative, if every measured squared stretch is in error by at most the same amount, the fitted squared stretch changes by at most 1.441 times that amount (the largest total absolute weight over directions).
Second, by Weyl's inequality and the Davis-Kahan theorem \citep{yu2015davis}, the relative error of anisotropy scales with the tensor error divided by the least squared stretch, and the bearing error with the tensor error divided by the difference between the two principal squared stretches.
Anisotropy is therefore least reliable at strongly anisotropic sites, and bearing is poorly determined near isotropy.
For encoders that are three times continuously differentiable, the leading departure from linearity changes sign between opposite probes and enters only the first-harmonic coefficients and the residual, so the remaining error of the tensor is of order $r^2$, and halving the radius reduces it about fourfold, as the double-precision CSP sweep shows (Section~\ref{supp:csp}).

A close fit does not, however, guarantee accuracy, since part of this $r^2$ term is itself elliptical and biases the tensor without appearing in the residual.
As a specific example, the encoder $\mathbf{f}(x,y)=(\sin2\pi Ax,\cos2\pi Ax,\sin2\pi y,\cos2\pi y)$ has limiting anisotropy $A$; fitted with 32 bearings at $r=1/(16A)$, it gives $\nu\le0.0023$ for $A$ from 2 to 20, yet its fitted anisotropy is 9.235 instead of 10 at $A=10$, and the error reaches 22\% at $A=20$.

\paragraph{Rounding and non-smooth encoders} These results assume exact evaluation of an encoder that is smooth over the probe disc.
Rounding sets a lower limit on useful radii: an absolute feature error of at most $\varepsilon_f$ changes each finite stretch by up to $2\varepsilon_f/r$, so below a certain radius, which is smaller at higher numerical precision, this term dominates and doubles with each halving of $r$ (Section~\ref{supp:csp}; Figure~\ref{fig:radius-screening}).
Smoothness over the disc holds for the SIREN and GELU networks of SatCLIP, Climplicit and CSP-fMoW wherever the input coordinates are continuous (and, for unit-normalized outputs, the feature norm is nonzero), but holds for the ReLU-type networks of GeoCLIP, SINR and CSP-iNat only within a single activation region; for a disc that crosses a region boundary, the error need not shrink with the radius (Sections~\ref{supp:csp} and~\ref{supp:relu}).

The scalings above follow from a Taylor expansion of the encoder over the probe disc and standard matrix perturbation bounds.
Since bounds on the derivatives of the encoders are not estimated here, we instead assess accuracy empirically, against analytic references and model Jacobians (main Section~\ref{sec:validation}) and by radius refinement (Section~\ref{supp:radius}).

\subsection{Geographic reference formulas}\label{app:closed}

\begin{figure}[htbp]
\centering\includegraphics[width=\textwidth]{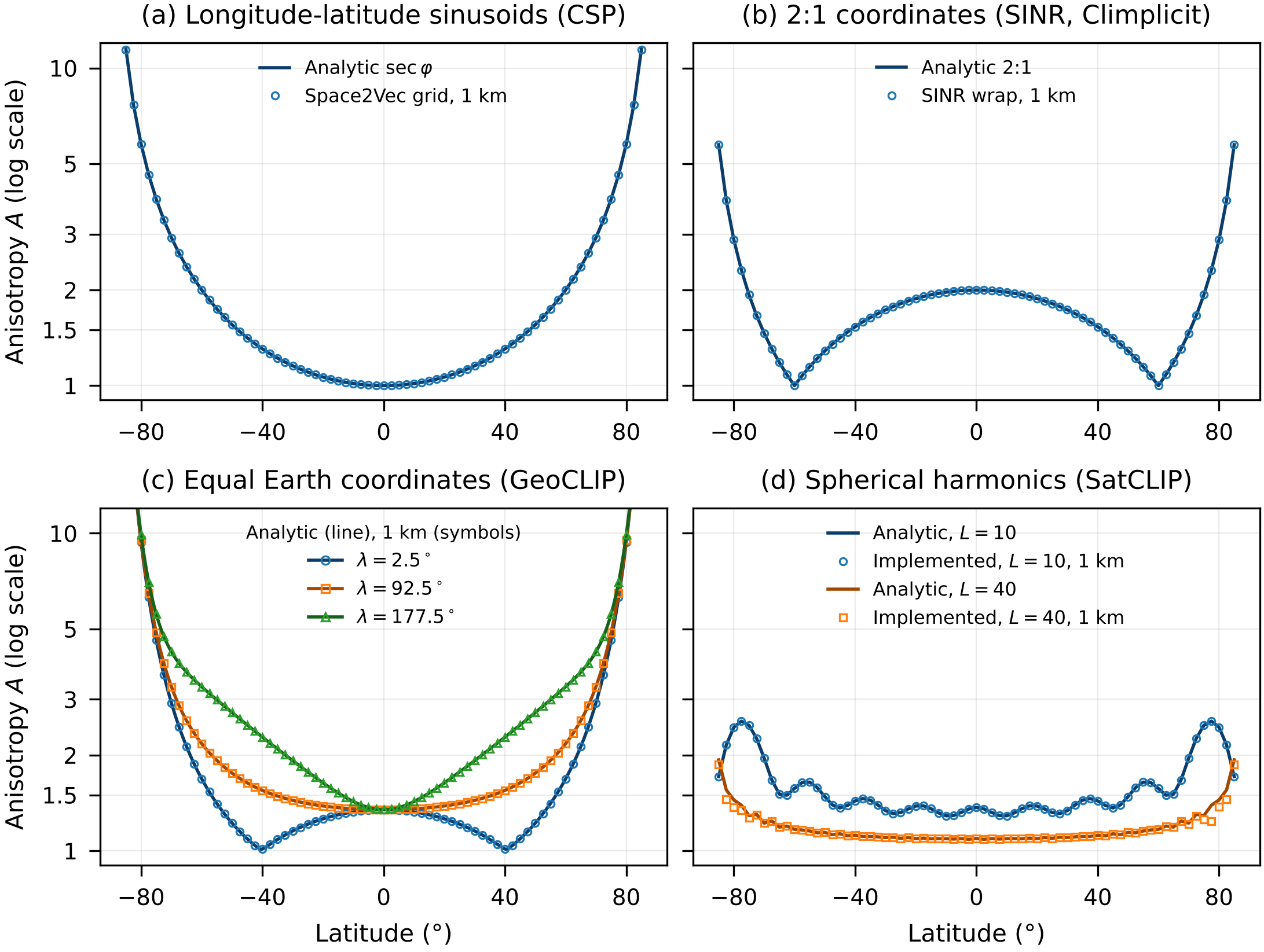}
\caption{Validation against known geographic input geometry. Lines show analytic infinitesimal anisotropy (main Table~\ref{tab:refs}), and symbols show median anisotropy fitted with main Equation~\ref{eq:profile-fit} along latitude circles at the probe radii given in the legends, on a shared logarithmic axis. (a) Longitude-latitude sinusoids (CSP). (b) 2:1 coordinates (SINR, Climplicit). (c) Equal Earth (GeoCLIP) along three meridians; symbols are single sites rather than latitude medians, because this reference varies with longitude. (d) SatCLIP harmonics with the zonal components scaled by $\pi$, as implemented; $L$ retains degrees $0,\ldots,L-1$, and departures at $L=40$ reflect numerical error in the implemented expressions. All comparisons concern input encodings before the learned network.}
\label{fig:validation}
\end{figure}

The derivations below follow the recipe in the note to main Table~\ref{tab:refs}; the expressions require positive stretches and therefore exclude the poles. Figure~\ref{fig:validation} compares the resulting analytic anisotropy with the estimates fitted directly to each encoding, as summarized in main Section~\ref{sec:closed-form}.

\paragraph{Equally scaled pairs (CSP)} For $(\sin\lambda,\cos\lambda,\sin\varphi,\cos\varphi)$, or any set of pairs with identical frequencies for both coordinates, the stretches are proportional to $1/R$ and $1/(R\cos\varphi)$, so $A=\sec\varphi$ with an east-west axis.

\paragraph{2:1 coordinates (SINR, Climplicit)} The wrap $(\sin\lambda,\cos\lambda,\sin2\varphi,\cos2\varphi)$ \citep[Equation~2]{cole2023sinr} has stretches $2/R$ and $1/(R\cos\varphi)$, because latitude is divided by $90^\circ$ and longitude by $180^\circ$, and the coordinates $(\lambda/\pi,2\varphi/\pi)$ of Climplicit give the same ratio.

\paragraph{Spherical harmonics (SatCLIP)} For orthonormal real harmonics of degrees $0\le l<L$, both squared stretches equal $S_{\rm iso}/R^2$ with $S_{\rm iso}=\sum_{l<L}(2l+1)l(l+1)/(8\pi)$. Multiplying the zonal components ($m=0$) by $c$ adds $(c^2-1)Z/R^2$ northward, where $Z(\vartheta)=\sum_{l<L}\frac{2l+1}{4\pi}[\sin\vartheta P_l'(\cos\vartheta)]^2$, $\vartheta=\pi/2-\varphi$ is colatitude and $P_l'$ is the derivative of the Legendre polynomial of degree $l$, so the ratio of northward to eastward squared stretch is $q_{\rm SH}=1+(c^2-1)Z/S_{\rm iso}$; evaluating $P_l'$ at high degree requires stable recurrences.

\paragraph{Equal Earth (GeoCLIP)} The reference is the Tissot indicatrix of the Equal Earth projection, computed from the analytic partial derivatives of the projection equations; its principal axes are generally oblique.

\section{Numerical checks of the estimator}\label{supp:radius}

This section reports three numerical checks of the estimator, namely radius refinement, sensitivity to the number of bearings and the variation of finite stretch with probe radius.

\paragraph{Radius refinement} When a derivative bound is unavailable (Section~\ref{supp:bound}), radius refinement provides an empirical convergence diagnostic. Specifically, the radius is halved until the operator norm of the difference between successive fitted tensors, divided by the finer tensor's minor eigenvalue, is at most 0.005 in two consecutive comparisons, with no fallback in either, although agreement across radii does not exclude a shared bias and is therefore not a certificate of accuracy. With exact evaluation, the rule converges in all 119 local sinusoidal cases (seven anisotropies from $A=1.5$ to 50 and 17 orientations) and in five spherical cases at latitudes from $60^\circ$ to $89^\circ$, with relative anisotropy errors below $10^{-4}$. On the released checkpoints, the rule was met in 34 of 50 cases selected to be difficult (high anisotropy, initial fallbacks and proximity to the antimeridian), and the largest accepted GeoCLIP tensor error against the Jacobian was 0.78\%, above the stopping threshold; the CSP failures are examined in Section~\ref{supp:csp}.

\paragraph{Number of bearings} To assess the sensitivity to the number of bearings, profiles measured at 64 bearings at 200 sites per checkpoint within $60^\circ$ (seed 20261016), each at its primary radius, were fitted in full and after subsampling to 32 and 16 bearings. Relative to 32 bearings, the 95th-percentile change in anisotropy is largest for GeoCLIP (0.15\% with 16 bearings and 0.03\% with 64) and at most 0.09\% for the other checkpoints, and the 95th-percentile change in principal bearing (sites with $A>1.05$) is at most $0.15^\circ$ for every checkpoint, which indicates that, within the tested range, the number of bearings has little effect on the estimates and 32 bearings resolve the second-harmonic profile with a wide margin.

\begin{figure}[htbp]\centering\includegraphics[width=\textwidth]{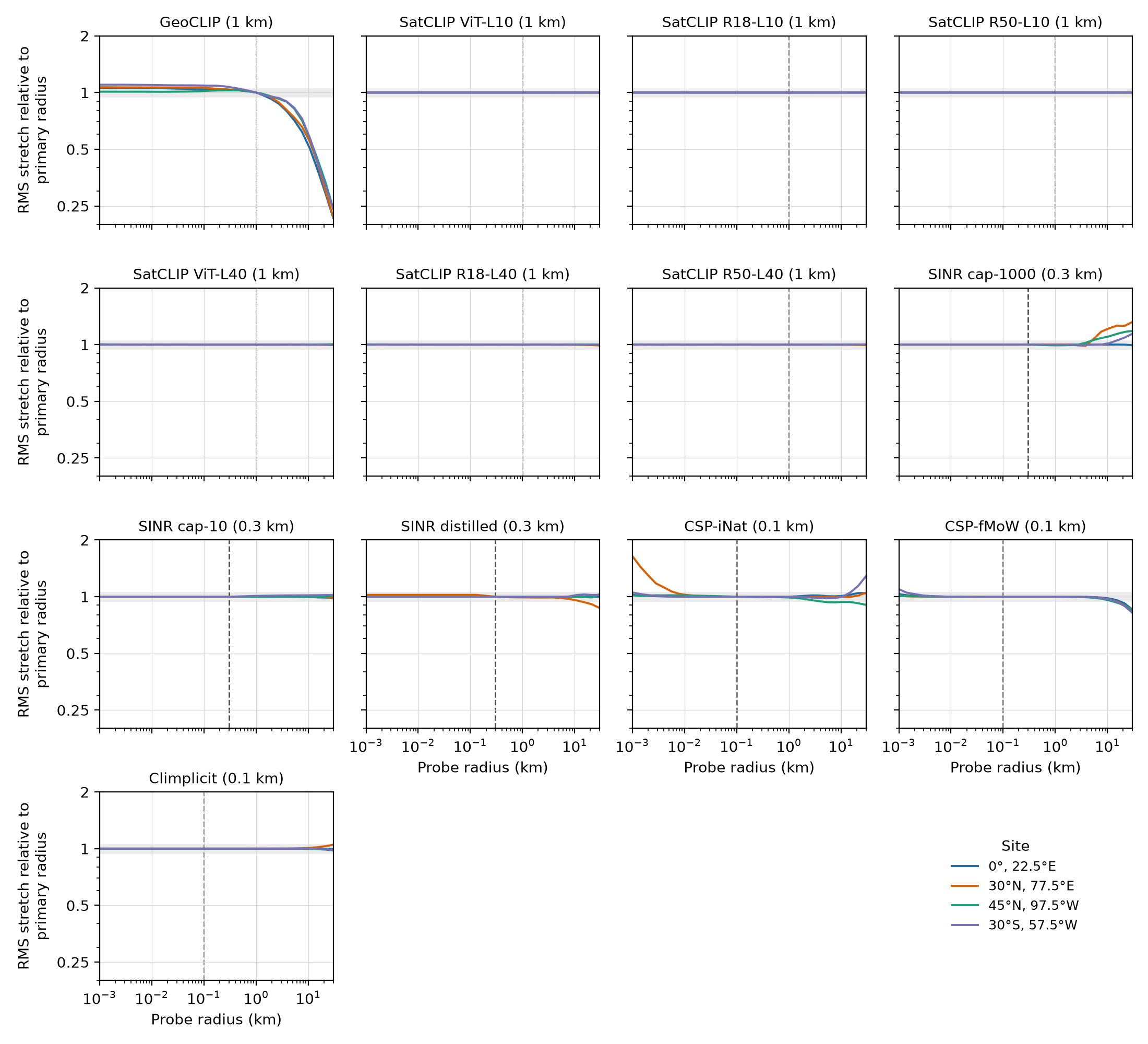}
\caption{Variation of finite stretch with probe radius for the thirteen released checkpoints at four fixed-grid sites. Each curve is the root-mean-square of the finite stretch $s_r$ (Equation~\ref{eq:finite-stretch}) over 32 bearings, divided by its value at the checkpoint's primary radius (dashed line, given in each title). The gray band marks $\pm5\%$.}
\label{fig:radius-screening}
\end{figure}

\paragraph{Radius screening} To confirm that the primary radii lie where the first-order approximation holds (main Section~\ref{sec:validity}), Figure~\ref{fig:radius-screening} shows how each checkpoint's stretch varies with radius. At four fixed-grid land sites, each released checkpoint was evaluated on 32-bearing rings at radii from 1\,m to 30\,km, and the root-mean-square of the finite stretch $s_r(\bp,\theta_j)$ (main Equation~\ref{eq:finite-stretch}) over the 32 bearings is shown relative to its value at the primary radius; these curves were computed after the primary radii had been chosen. A flat curve indicates that feature differences grow in proportion to distance, a falling curve indicates saturation, oscillation or many activation regions within the disc, and a rising curve at small $r$ reflects rounding; flatness is necessary but not sufficient for an accurate tensor (main Section~\ref{sec:validation}).

The curves are flat around the primary radius for SatCLIP (to 30\,km), SINR (to a few kilometers), Climplicit (to about 10\,km) and CSP (from about 10\,m to about 1\,km; below 10\,m, single-precision rounding raises them). GeoCLIP's stretch falls to about half its small-radius value at 10\,km, so its 1\,km primary radius lies at the start of the departure from linearity (main Sections~\ref{sec:validity} and~\ref{sec:jacobian}).

\section{Geographic sampling and coordinate boundaries}\label{supp:sampling}

\paragraph{Displaced grids} The eight finer fields summarized on displaced grids (main Section~\ref{sec:protocol}) are the three SatCLIP L40 checkpoints, SINR cap-1000 and distilled, both CSP checkpoints and Climplicit. Within $60^\circ$, their log anisotropy has lag-one correlations between adjacent fixed-grid nodes of 0.18--0.47 along longitude and 0.27--0.54 along latitude, against 0.71--0.85 and 0.79--0.87 for the five smoother fields, which indicates that the fixed grid poorly resolves their spatial variation. To construct the displaced grids, each base-grid node is moved uniformly within its latitude-longitude cell, independently for two seeds, and the randomized coordinates determine the weights, latitude masks and block membership; SatCLIP R18-L40 and R50-L40 use the same randomized coordinates as the ViT profiles, which supports comparison across these backbones. For the six finer fields compared across grid phases, $r_A$ on the fixed grid and the two randomized grids differs by at most 0.03, except for CSP-iNat ($-0.14$ to $-0.05$), and $C_\theta$ on the fixed grid is within 0.01 of the pooled value, except for CSP-iNat (0.24 against 0.19) and CSP-fMoW (0.22 against 0.14), so additional sampling may be needed for finer spatial structure.

\paragraph{Block resampling} Block resampling provides the 95\% intervals in main Table~\ref{tab:enc} and describes how the summaries of each checkpoint depend on the geographic composition of its sample. Because the anisotropies of neighboring sites are correlated, sites within $60^\circ$ are resampled in latitude-longitude blocks rather than individually, with blocks of $30^\circ\times60^\circ$ for the five smoother fields and $15^\circ\times30^\circ$ for the eight finer fields (see Displaced grids). Each replicate draws as many blocks as the sample contains, with replacement, and recomputes $r_A$ and $C_\theta$ after multiplying the area weight of each site by the number of times its block is drawn. The interval is then given by the 2.5th and 97.5th percentiles over 1,000 replicates (random seed 16). These intervals are descriptive rather than inferential, as neighboring blocks and checkpoints with related training data do not form independent replications, and they are therefore not used to declare significance from the exclusion of zero. In addition, the larger blocks yield only about 30 distinct blocks within $60^\circ$, so the intervals are coarse and can be asymmetric, and because the block sizes differ, the intervals of the smoother and finer fields are not directly comparable.

\paragraph{Admissible bearings} To avoid assigning a preferred direction to an effectively isotropic response, $C_\theta$ excludes sites where either anisotropy is at most $1+10^{-6}$. To assess the sensitivity of the result to this restriction, we also apply fixed thresholds of 1.01 and 1.05, set before agreement was compared. For every checkpoint except CSP, $C_\theta$ changes by at most 0.01 and at least 97\% of the area is retained. For CSP, $C_\theta$ rises from 0.19 to 0.28 (CSP-iNat) and from 0.14 to 0.26 (CSP-fMoW) while the retained area falls to 65\%, because the stricter thresholds remove the equatorial band where its $\sec\varphi$ input reference is nearly isotropic ($|\varphi|<17.8^\circ$ at 1.05).
\paragraph{Land-only summaries} Because training observations are predominantly on land, Table~\ref{tab:land} repeats the Table~\ref{tab:enc} statistics on land sites only. GeoCLIP's association with Equal Earth is unchanged, whereas several latitude-only associations change (for example, Climplicit's $r_A$ falls from 0.19 to 0.02 and SINR cap-10's strengthens from $-0.19$ to $-0.44$), which indicates that they depend partly on the behavior of the encoders over the ocean.

\begin{figure}[htbp]\centering\includegraphics[width=\textwidth]{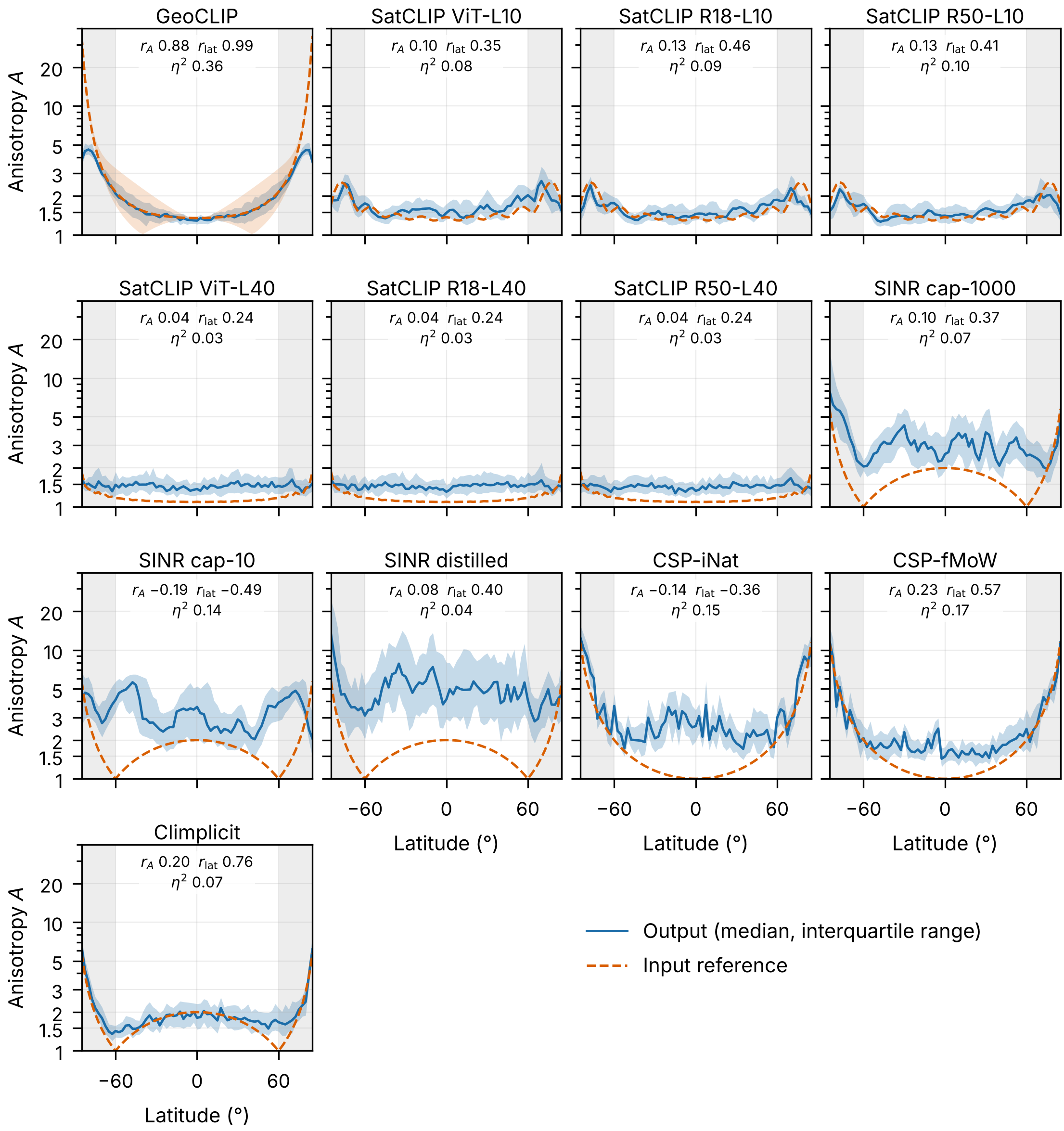}
\caption{Latitude profiles of output anisotropy (blue; median and interquartile range in each $2.5^\circ$ latitude row) and of the input reference (orange dashed; for GeoCLIP, the median and range across longitudes) on the fixed grid at the primary probe radii. Gray bands mark latitudes beyond $60^\circ$, which are excluded from $r_A$. Each panel gives $r_A$, the latitude-row correlation $r_{\rm lat}$ and the latitude share $\eta^2$ within $60^\circ$. For the eight checkpoints summarized on randomized sites in main Table~\ref{tab:enc}, the fixed-grid $r_A$ shown here can differ from the tabulated value.}
\label{fig:latitude}
\end{figure}

\begin{table}[htbp]\centering\small
\caption{Table~\ref{tab:enc} statistics over all sites and over land sites only. Land is defined by the 0.5$^\circ$ land mask used for the map coastlines; land covers 29\% of the area within $85^\circ$. Same samples, weights and exclusions as Table~\ref{tab:enc}.}
\label{tab:land}
\begin{tabular}{@{}lrrrrrr@{}}\toprule
&\multicolumn{2}{c}{Median $A$}&\multicolumn{2}{c}{$r_A$}&\multicolumn{2}{c}{$C_\theta$}\\
Checkpoint&All&Land&All&Land&All&Land\\\midrule
GeoCLIP&1.46&1.47&0.88&0.86&0.88&0.87\\
SatCLIP ViT-L10&1.57&1.84&0.10&0.07&0.49&0.42\\
SatCLIP R18-L10&1.50&1.72&0.13&0.07&0.49&0.39\\
SatCLIP R50-L10&1.50&1.67&0.13&0.11&0.55&0.44\\
SatCLIP ViT-L40&1.44&1.56&0.02&$-$0.02&$-$0.01&0.12\\
SatCLIP R18-L40&1.44&1.59&0.03&$-$0.01&0.02&0.14\\
SatCLIP R50-L40&1.43&1.58&0.04&0.01&0.02&0.14\\
SINR cap-1000&2.89&2.73&0.09&0.01&0.38&0.32\\
SINR cap-10&2.99&2.50&$-$0.19&$-$0.44&0.44&0.61\\
SINR distilled&4.88&3.71&0.06&0.06&0.25&0.24\\
CSP-iNat&2.44&2.63&$-$0.07&$-$0.15&0.19&0.18\\
CSP-fMoW&1.81&1.87&0.23&0.31&0.14&0.14\\
Climplicit&1.77&1.78&0.19&0.02&0.56&0.35\\
\bottomrule\end{tabular}
\end{table}
\paragraph{Latitude profiles} Figure~\ref{fig:latitude} compares the latitude profiles of output anisotropy with those of the input references on the fixed grid. For a reference that varies only with latitude, $r_A=r_{\rm lat}\sqrt{\eta^2}$, where $r_{\rm lat}$ is the area-weighted correlation between the latitude-row means of log output anisotropy and the reference profile, and $\eta^2$ is the fraction of the variance of log output anisotropy within $60^\circ$ explained by latitude. Across the twelve latitude-only references, $\eta^2$ ranges from 0.03 to 0.17, which limits $|r_A|$ to at most $\sqrt{0.17}\approx0.41$, so even closely matching profiles give small $r_A$; for example, $r_{\rm lat}$ is 0.76 for Climplicit and 0.57 for CSP-fMoW, whereas the profiles of SINR cap-10 and CSP-iNat run opposite to their references ($r_{\rm lat}$ of $-0.49$ and $-0.36$). A matching profile does not, however, identify the input reference as its source, since latitudinal patterns in the training data could produce a similar profile. GeoCLIP's reference also varies with longitude, so its two terms are shown for comparison only.

\paragraph{Coordinate boundaries} Table~\ref{tab:boundary} gives the per-checkpoint results of the boundary tests in main Section~\ref{sec:boundaries}, with all outputs L2-normalized for this comparison, which provides a bounded scale without changing the raw-feature quantity used in the main SINR and Climplicit analysis.
\begin{table}[htbp]\centering\small
\caption{Coordinate-boundary checks on released outputs. Seam distances are ranges across the three test latitudes at the smallest longitude offset; pole distances compare two descriptions of one point.}
\label{tab:boundary}
\begin{tabular}{@{}lrr@{}}\toprule
Checkpoint&Seam distance&Same-pole distance\\\midrule
GeoCLIP&1.04--1.30&1.09\\
SINR cap-1000&$2.92\times10^{-8}$--$3.64\times10^{-8}$&0.497\\
SINR cap-10&$4.62\times10^{-9}$--$8.31\times10^{-9}$&0.178\\
SINR distilled&$1.79\times10^{-8}$--$4.14\times10^{-8}$&0.336\\
SatCLIP ViT-L10&$1.25\times10^{-8}$--$8.83\times10^{-8}$&0\\
SatCLIP ViT-L40&$2.05\times10^{-7}$--$6.95\times10^{-7}$&0\\
SatCLIP R18-L10&$1.71\times10^{-8}$--$9.54\times10^{-8}$&0\\
SatCLIP R18-L40&$2.16\times10^{-7}$--$7.71\times10^{-7}$&0\\
SatCLIP R50-L10&$1.64\times10^{-8}$--$7.96\times10^{-8}$&0\\
SatCLIP R50-L40&$2.18\times10^{-7}$--$7.47\times10^{-7}$&0\\
CSP-iNat&0.029--0.162&0.0523\\
CSP-fMoW&0.0616--0.239&0.238\\
Climplicit&0.795--1.22&1.15\\
\bottomrule\end{tabular}
\par\smallskip\parbox{.97\textwidth}{\footnotesize A vanishing seam distance does not by itself certify differentiability.}
\end{table}

For GeoCLIP, CSP and Climplicit, a probe ring is excluded from the summaries if its disc intersects the antimeridian. For a center in the adjacent longitude half-plane, the clearance is $R\arcsin(|\cos\varphi\sin(\pi-|\lambda|)|)$, and near the seam, radius refinement (Section~\ref{supp:radius}) halves the starting radius until it is smaller than this clearance. For SINR and SatCLIP, whose outputs are continuous across the seam, rings that cross it are retained. Fixed-grid centers lie beyond every probe radius from the antimeridian, so exclusions arise only on the randomized grids, where they remove one ring each for CSP-iNat, CSP-fMoW and Climplicit.

\section{Jacobian validation and refinement limits}\label{supp:jacobian}

To assess the finite-radius estimates independently, we use Jacobians computed by automatic differentiation (main Section~\ref{sec:jacobian}) as a reference for seven checkpoints at 2,340 sites through $80^\circ$. At each site, the error of the fitted anisotropy $A_r$ at the primary radius is measured relative to the Jacobian anisotropy $A_J$ as $|A_r/A_J-1|$, and Table~\ref{tab:tails} summarizes this error for five of the checkpoints (SINR cap-10 and distilled are covered in Section~\ref{supp:relu}). Since small median errors can coexist with substantial errors at strongly anisotropic locations, the table also reports the error in the upper decile of $A_J$ and the number of fallbacks. CSP is assessed separately in Section~\ref{supp:csp}, since its released code evaluates the network in single precision.

\begin{table}[htbp]\centering\small
\caption{Error of the finite-radius anisotropy against the Jacobian, $|A_r/A_J-1|$, at each checkpoint's primary radius. Errors are area-weighted and in percent; High-$A$ denotes the upper decile of Jacobian anisotropy, and Fallbacks gives the number of the 2,340 sites that required the fallback.}
\label{tab:tails}
\begin{tabular}{@{}lrrrr@{}}
\toprule
Encoder&Median&95th percentile&High-$A$: 95th&Fallbacks\\\midrule
GeoCLIP&4.43&16.7&52.0&19\\
SatCLIP ViT-L10&$3.44\times10^{-5}$&$1.66\times10^{-4}$&$3.70\times10^{-4}$&0\\
SatCLIP ViT-L40&$8.58\times10^{-4}$&0.240&0.497&3\\
SINR cap-1000&$4.21\times10^{-7}$&1.15&16.1&10\\
Climplicit&$5.77\times10^{-5}$&$6.45\times10^{-4}$&0.00407&0\\
\bottomrule\end{tabular}

\end{table}

\subsection{Double-precision check of CSP}\label{supp:csp}
\paragraph{Precision at the primary radius} The released CSP code computes its sinusoidal input features in double precision from coordinates in degrees, then casts them to single precision before the feed-forward network. To separate encoder geometry from rounding, we evaluated both unsupervised CSP checkpoints with the same weights entirely in double precision, verified that the double-precision input features match the released ones to $10^{-12}$, and computed double-precision automatic Jacobians of the unit-normalized output in the north-east frame. The comparison uses 300 sites drawn uniformly by area within $60^\circ$ (seed 20261015), excluding the $1^\circ$ nearest the antimeridian, with the 0.1\,km primary radius and 32 bearings. Since the sites are already drawn uniformly by area, the error $|A_r/A_J-1|$ is summarized without area weights (Table~\ref{tab:cspprecision}).
\begin{table}[htbp]\centering\small
\caption{CSP anisotropy error against double-precision Jacobians, $|A_r/A_J-1|$ in percent, at 300 sites within $60^\circ$ and the 0.1\,km primary radius. Summaries are unweighted, and Median $A_J$ is the median Jacobian anisotropy itself. High-$A$ denotes the upper decile of Jacobian anisotropy. Crossing is the fraction of probe discs in which at least one probe point has a different activation pattern from the center.}
\label{tab:cspprecision}
\begin{adjustbox}{max width=\textwidth}
\begin{tabular}{@{}lrrrrrrrr@{}}\toprule
&&\multicolumn{3}{c}{Single precision (released)}&\multicolumn{3}{c}{Double precision}&\\
Checkpoint&Median $A_J$&Median&95th&High-$A$ 95th&Median&95th&High-$A$ 95th&Crossing\\\midrule
CSP-iNat (leaky ReLU)&2.44&0.022&0.97&2.78&0.0017&0.97&2.72&21\%\\
CSP-fMoW (GELU)&1.68&0.0053&0.021&0.031&0.0006&0.0041&0.0085&---\\
\bottomrule\end{tabular}
\end{adjustbox}
\par\smallskip\parbox{.97\textwidth}{\footnotesize No site required a fallback in either precision. Crossing is undefined for the smooth GELU activation.}
\end{table}
As Table~\ref{tab:cspprecision} shows, single-precision rounding contributes little at the primary radius; for CSP-fMoW, the 95th-percentile error is below 0.03\% in both precisions. For CSP-iNat, the released and double-precision errors have nearly the same upper tails, which therefore arise from nonlinearity within the probe disc rather than from precision. This tail coincides with discs that cross activation boundaries (Section~\ref{supp:bound}): all sites in the upper decile of error are crossing discs, and the median double-precision error is 0.24\% for the 62 crossing discs against 0.001\% for the 238 others.

\paragraph{Refinement failures} Radius refinement, however, fails for a different reason: with the released implementation, among the sites selected for difficulty (Section~\ref{supp:radius}), the refinement criterion is met at none of eight CSP-iNat sites and at only three of six CSP-fMoW sites. At $(55^\circ\mathrm{N},112.5^\circ\mathrm{E})$, one of these CSP-iNat sites, halving the radius repeatedly from 100\,m to 0.8\,m raises the single-precision anisotropy error to 22\%, whereas the double-precision error falls by a factor of about four per halving; the other swept sites (one more for CSP-iNat and two for CSP-fMoW) behave in the same way. Single-precision rounding, which grows as the radius shrinks, therefore prevents the successive-change criterion from being met, whereas in double precision the estimates converge to the Jacobian value. At the remaining refinement sites, the two precisions agree at the 0.1\,km radius but differ from the Jacobian, by 2.4--12.6\% at four CSP-iNat sites whose discs cross activation boundaries and by 0.3--0.4\% at two CSP-fMoW sites with anisotropy above 11.

\subsection{Activation-region crossings}\label{supp:relu}
GeoCLIP and SINR use ReLU activations, so their finite-radius error depends on whether a probe disc crosses an activation-region boundary (Section~\ref{supp:bound}). To detect such crossings, we compared the activation pattern of every hidden unit at the 32 probe points with that at the center, at each of the 2,340 Jacobian sites and with the same probe rings as the finite-radius estimates; Table~\ref{tab:relu} summarizes the results discussed in main Section~\ref{sec:jacobian}, with errors $|A_r/A_J-1|$ as in Table~\ref{tab:tails}. For SINR, every site with error above 0.1\% has a detected crossing, although the test, which checks only the sampled probe points, does not exclude other contributions to the heavy tails. For GeoCLIP, every disc crosses at both 1 and 0.3\,km in its $\sigma=16$ and $\sigma=256$ Fourier-feature branches, and 64\% (1\,km) or 27\% (0.3\,km) also cross in its $\sigma=1$ branch, so GeoCLIP is never within a single activation region at these radii. The same check for CSP-iNat is reported in Section~\ref{supp:csp}.

\begin{table}[htbp]\centering\small
\caption{Activation-region crossings at the 2,340 Jacobian sites. A probe disc crosses when any probe point has a different ReLU activation pattern from the center. Errors are area-weighted $|A_r/A_J-1|$ in percent.}
\label{tab:relu}
\begin{adjustbox}{max width=\textwidth}
\begin{tabular}{@{}lrrrrr@{}}\toprule
&&\multicolumn{2}{c}{Median error (\%)}&\multicolumn{2}{c}{95th percentile (\%)}\\
Checkpoint, radius&Crossing discs&Within&Crossing&Within&Crossing\\\midrule
SINR cap-1000, 0.3 km&12.2\%&$3.0\times10^{-7}$&0.65&$4.7\times10^{-6}$&16.7\\
SINR cap-10, 0.3 km&11.6\%&$4.6\times10^{-7}$&0.12&$1.9\times10^{-5}$&1.79\\
SINR distilled, 0.3 km&5.8\%&$7.4\times10^{-7}$&5.77&$1.4\times10^{-5}$&80.0\\
\midrule
&Median switching units&\multicolumn{2}{c}{Lowest / highest switching tercile}&\multicolumn{2}{c}{Lowest / highest switching tercile}\\
GeoCLIP, 1 km&354&\multicolumn{2}{c}{3.45 / 9.04}&\multicolumn{2}{c}{9.96 / 39.7}\\
GeoCLIP, 0.3 km&108&\multicolumn{2}{c}{1.82 / 3.24}&\multicolumn{2}{c}{5.24 / 18.1}\\
\bottomrule\end{tabular}
\end{adjustbox}
\par\smallskip\parbox{.97\textwidth}{\footnotesize SINR Jacobians are double-precision automatic derivatives; for cap-1000 they reproduce the stored values to $3\times10^{-14}$. Every GeoCLIP disc crosses at least one of its 9,216 units at both radii; the rows therefore compare terciles of the number of units whose pattern changes within the disc (Spearman correlation with error 0.53 at 1 km, 0.39 at 0.3 km).}
\end{table}

\section{GeoCLIP and the Equal Earth reference}\label{supp:geoclip}

\paragraph{Fourier-feature anisotropy} To quantify the anisotropy attributed in main Section~\ref{sec:geoclip} to the random Fourier features, we note that for features $(\cos \bB\bx,\sin \bB\bx)$ with frequency matrix $\bB$, the input metric in projected coordinates is $\bB^\top \bB$, whose anisotropy does not depend on the frequency scale. For $\bB$ with 256 rows of independent standard Gaussian entries, the anisotropy has median 1.08 and 95th percentile 1.17 over 20,000 draws; this small departure from isotropy arises from the finite sample of isotropically distributed frequencies.

\paragraph{Comparison with Jacobians} Main Section~\ref{sec:geoclip} also reports that GeoCLIP's association with Equal Earth is nearly unchanged when the tensors are computed from Jacobians; on the 1,764 Jacobian sites within $60^\circ$, $r_A$ with Equal Earth is 0.874 for the finite fit and 0.869 for the Jacobian. The correlation $r_A$ compares scalar anisotropy across sites, whereas a pointwise comparison of tensors also accounts for orientation. For positive tensors $\bG_{\rm in}$ and $\bG_{\rm out}$ at a site, let $\mu_{\max}\ge\mu_{\min}>0$ be the eigenvalues of $\bG_{\rm in}^{-1}\bG_{\rm out}$. The shape discrepancy
\[
D=\sqrt{\mu_{\max}/\mu_{\min}}
\]
is the anisotropy of $\bG_{\rm out}$ measured relative to $\bG_{\rm in}$. It equals one when the two tensors differ only by a scalar factor, and when either tensor is isotropic it reduces to the anisotropy of the other, so an isotropic output gives the anisotropy of Equal Earth itself as a baseline. Table~\ref{tab:shape} compares GeoCLIP's finite-fit and Jacobian tensors with the analytic Equal Earth tensor in the north-east frame on these sites. Both lie much closer to Equal Earth than this baseline, and they differ from each other by less than either differs from Equal Earth, so the remaining departure from Equal Earth is not mainly estimation error. These comparisons measure agreement in shape; they do not show that the Equal Earth input causes it, or how much of the input geometry survives training.
\begin{table}[htbp]\centering\small
\caption{Area-weighted shape discrepancy $D$ for GeoCLIP on the same 1,764 Jacobian sites within $60^\circ$. $D=1$ denotes identical shape; the isotropic row is the anisotropy of Equal Earth itself.}
\label{tab:shape}
\begin{tabular}{@{}lrr@{}}\toprule
Comparison&Median $D$&95th percentile\\\midrule
Finite fit / Equal Earth&1.147&1.331\\
Jacobian / Equal Earth&1.162&1.375\\
Finite fit / Jacobian&1.067&1.152\\
Isotropic / Equal Earth&1.380&2.320\\
\bottomrule\end{tabular}

\end{table}

\section{SatCLIP implementation and input-rescaling comparison}\label{supp:satclip}
Both issues described in this section, the scaling of the zonal components and the numerical errors of high-degree harmonics near the poles, arise in the spherical-harmonic positional encoding of SatCLIP, that is, in the input presented to the network, rather than in the learned network itself. They therefore affect the closed-form check in main Section~\ref{sec:closed-form} and the input reference in main Section~\ref{sec:protocol}, whereas their effect on the released output geometry is examined separately in Section~\ref{supp:satclip-rescale}.

\paragraph{Zonal scaling of the input} The analysis uses SatCLIP commit \texttt{1ead106}, obtained on 12 June 2026, whose harmonic table originates in the November 2023 code, and the measurements reported here are therefore specific to this code version and its released checkpoints. In this code, the input geometry is set by the scaling of the zonal components (harmonic order $m=0$). Specifically, the implemented table uses an expression equivalent to $\sqrt{(2l+1)\pi/4}$, whereas orthonormal harmonics require $\sqrt{(2l+1)/(4\pi)}$, giving a ratio of $\pi$. For degree limit $L=10$, the resulting input anisotropy ranges from approximately 1.29 to 1.65 within $60^\circ$ latitude and reaches 2.56 at $85^\circ$, whereas for $L=40$ it ranges from 1.09 to 1.18 within $60^\circ$ (see Section~\ref{app:closed} for the general scaling formula). As the compensated control in Section~\ref{supp:satclip-rescale} shows, this $\pi$ scaling can be absorbed into the first-layer weights.

\paragraph{Numerical cancellation in the input} At high degrees, the evaluation of the harmonics introduces a further discrepancy. The code computes each harmonic from an expanded polynomial in $\cos\theta$, whose coefficients alternate in sign and, at degree 39, reach about $10^{14}$ while being stored with about 15 significant digits. Near the poles, where $|\cos\theta|$ approaches one, the terms of this polynomial remain large and must cancel to a value of order one, that is, a small value is obtained as the difference of large, nearly equal terms. Since double precision retains only about 16 significant digits, terms that are about $10^{15}$ times larger than the result leave at most one or two correct digits, so small relative errors in the stored coefficients and in the arithmetic become large absolute errors in the basis functions; at $77.5^\circ$N, for example, the largest term of the zonal harmonic of degree 39 is about $6\times10^{13}$, and its computed value is $-0.40$ instead of $-0.077$. Toward the equator, $|\cos\theta|$ decreases, the high powers become negligible and the cancellation disappears. Against SciPy references in double precision, the largest absolute errors in individual basis functions grow with degree, from about $3\times10^{-3}$ at degree 35 to 0.1--0.5 at degree 39 between $70^\circ$ and $85^\circ$, so at high latitudes the L40 input field computed by the code departs from its analytic reference (main Section~\ref{sec:closed-form}), whereas the L10 input is unaffected. The errors are largest in the zonal and low-order functions, and the $\pi$ scaling of the zonal functions amplifies their effect on anisotropy; at $77.5^\circ$N, the fitted input anisotropy at 1\,km departs from its analytic value by 11\% with the scaling and by about 1\% without it. Evaluating the harmonics by recurrence, as in SciPy or the closed-form option of the SatCLIP code, avoids these large intermediate terms and hence the cancellation.

\begin{figure}[ht]\centering\includegraphics[width=\textwidth]{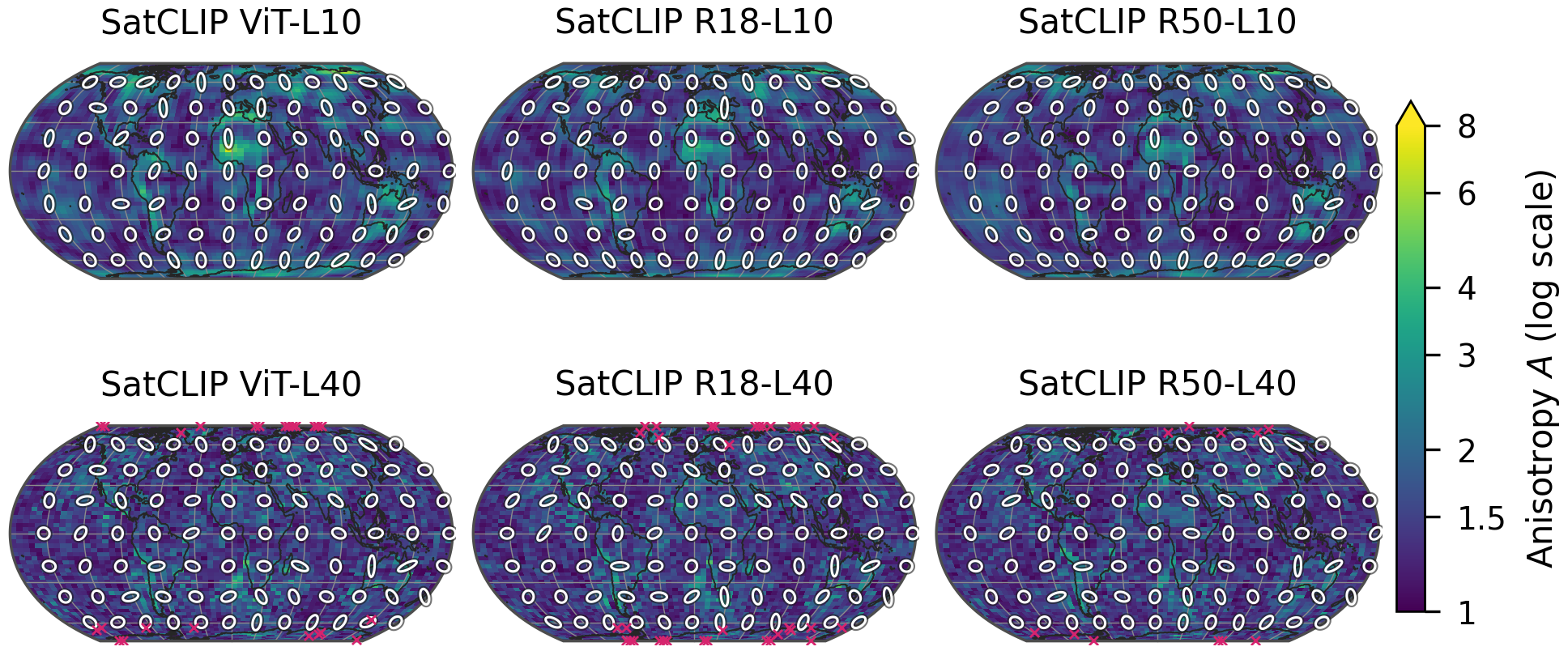}
\caption{SatCLIP fields on the fixed display grid in the Equal Earth projection, computed from the stored profiles with the tensor fit of main Equation~\ref{eq:profile-fit}. The numerical summaries in main Table~\ref{tab:enc} use randomized sites, not this display grid. Color shows anisotropy on the logarithmic scale shared with main Figures~\ref{fig:geoclip} and~\ref{fig:dense}; white ellipses within $60^\circ$ are drawn in the manner of Tissot's indicatrix, as in Figure~\ref{fig:geoclip}. Pink crosses mark sampled-minimum fallbacks. The checkpoint comparison confounds architecture and training effects.}
\label{fig:satclip-backbones}
\end{figure}

\subsection{Paired input rescaling of the released checkpoints}\label{supp:satclip-rescale}
To isolate the effect of scaling the zonal ($m=0$) components, we divide each $m=0$ harmonic channel by $\pi$ while keeping the learned weights fixed. It should be noted that this intervention changes the inputs presented to the pretrained network and is not equivalent to retraining with the corrected basis. In addition, we retain the expanded-polynomial evaluation, so the intervention changes the scaling but not the numerical errors described above.

The paired test covers all six SatCLIP checkpoints at 35 sites, namely seven latitudes from $-75^\circ$ to $75^\circ$ at $25^\circ$ intervals crossed with five longitudes from $-150^\circ$ to $150^\circ$ at $75^\circ$ intervals. Each site uses 32 bearings at radii of 1 and 0.5\,km and float64 evaluation; output profiles use the unit-normalized network outputs and input profiles the unnormalized harmonic features, and we report area-weighted medians on this selected grid.

\begin{table}[htbp]\centering\small
\caption{Paired SatCLIP input rescaling at 35 selected sites and 1\,km radius. Entries are area-weighted medians of fitted anisotropy; cosine is the area-weighted median cosine similarity between released and rescaled unit outputs at the site centers.}
\label{tab:satclipcorrection}
\begin{tabular}{@{}lrrrrr@{}}\toprule
&\multicolumn{2}{c}{Input $A$}&\multicolumn{2}{c}{Output $A$}&\\
Checkpoint&Released&Rescaled&Released&Rescaled&Cosine\\\midrule
ViT-L10&1.369&1.000&1.680&1.536&0.937\\
ViT-L40&1.095&1.000&1.397&1.432&0.979\\
R18-L10&1.369&1.000&1.575&1.417&0.928\\
R18-L40&1.095&1.000&1.421&1.424&0.974\\
R50-L10&1.369&1.000&1.509&1.367&0.927\\
R50-L40&1.095&1.000&1.513&1.511&0.975\\
\bottomrule\end{tabular}
\par\smallskip\parbox{.97\textwidth}{\footnotesize Medians include flagged fallback values. Released/rescaled fallback counts at 1 km, in row order: 0/0, 1/0, 0/0, 0/0, 0/0, 0/0.}
\end{table}
The rescaled L10 input has fitted anisotropy 1 to the displayed precision, whereas the rescaled L40 input has a maximum of $A=1.00277$ at 1\,km and $A=1.00828$ at 0.5\,km. This residual anisotropy grows as the radius shrinks, as expected for evaluation error (Section~\ref{supp:bound}), which confirms that dividing the zonal channels by $\pi$ leaves the numerical errors in place. Output anisotropy, however, remains above one after rescaling, with no consistent reduction across the L40 checkpoints.

In a separate control, we correct the harmonic basis and multiply the corresponding first-layer weight columns by $\pi$, which preserves the network's preactivation. Across the six checkpoints and 35 sites, unit outputs then differ from the released ones by at most $1.9\times10^{-16}$ and fitted anisotropies by a relative $3\times10^{-13}$. The released output geometry therefore does not depend on the basis convention, which affects only the input reference used for $r_A$ and $C_\theta$, the input-rescaled variants and possibly the training dynamics.

\FloatBarrier

\section{Checkpoint provenance and reproducibility}\label{supp:checkpoints}

Checkpoint labels follow the released models, in which CSP checkpoints are named by their image datasets, iNaturalist 2018 (iNat) and Functional Map of the World (fMoW); both checkpoints were trained without supervision. SatCLIP labels give the image backbone, ViT for a vision transformer and R18 and R50 for ResNet-18 and ResNet-50, together with the harmonic degree limit $L$. SINR checkpoints are labeled by the maximum number of training observations per species (cap-10, cap-1000) or as distilled from a model with environmental inputs. Feature dimensions are given in main Table~\ref{tab:encoders}, and measured vectors and probe radii in main Table~\ref{tab:enc}. Evaluation is in float64 except for GeoCLIP, which keeps its released float32 accumulation and output with float64 weights and inputs, and CSP, which keeps its released float32 forward pass with outputs stored as float64. Longitudes are wrapped to $[-180^\circ,180^\circ)$ before encoding; the seam and pole tests in Section~\ref{supp:sampling} assess coordinate equivalence.

Weights were obtained from the original releases: GeoCLIP from the \texttt{geoclip} 1.2.1 package (GitHub revision \texttt{a32dff0}; \citealp{cepeda2023geoclip}), SINR from Caltech Data record dk5g7-rhq64 \citep{cole2023sinr}, SatCLIP from the Microsoft Hugging Face repositories at pinned revisions \citep{klemmer2024satclip}, CSP from the authors' released model archive with code at revision \texttt{67f3407} \citep{mai2023csp}, and Climplicit from the Hugging Face repository \texttt{Jobedo/climplicit} at revision \texttt{fe34260} \citep{dollinger2025climplicit}. The weights remain subject to their source licenses: MIT for GeoCLIP and SatCLIP, CC-BY-4.0 for Climplicit, and the release terms for SINR and CSP.

The package supports two workflows: regenerating the tables and figures from saved measurements, and re-evaluating the checkpoints. Its documentation gives commands and checksums for each, and the package metadata record exact checkpoint identifiers and full hashes; model training and collection of the original training data are not reproduced.

\section{Independent neighborhood prediction and geographic reach}\label{supp:neighborhoods}

This section gives the details of the neighborhood comparison in main Section~\ref{sec:neighborhoods}, which covers the thirteen released checkpoints and the input-rescaled SatCLIP ViT-L10 and ViT-L40 variants of Section~\ref{supp:satclip}, with distances computed from all components of the measured feature vectors (main Section~\ref{sec:protocol}).

\paragraph{Query design} At each query, the 400 candidates are sampled uniformly by area with seed 20260918, and the same sampled bearings and relative distances are used across all models for disc radii of 0.3, 1, 3 and 6\,km (primary 1\,km). Tensors are fitted separately from 32-bearing profiles at radii 0.1, 0.3 and 1\,km (primary 0.3\,km), so the candidates do not enter the tensor fit. Fits requiring fallback are excluded from both geographic and tensor comparisons; at the primary fit radius, all models retain twenty queries except distilled SINR, which retains nineteen.

Actual feature distances are computed directly from differences of feature vectors, which avoids the cancellation that arises from subtracting cosine similarities near one. Rankings are compared for $k=10,20,40,80$ neighbors, with ties broken by stable candidate ordering, although no exact distance tie occurs at the primary selection boundary.

\paragraph{Reach ellipse} The reach ellipse in main Figure~\ref{fig:neighborhoods} is $\bdelta^\top\widehat{\bG}_r\bdelta=d_{40}^2$, where $\bdelta$ is a north-east geographic displacement in kilometers, $\widehat{\bG}_r$ is fitted from the separate 0.3\,km probes and $d_{40}$ is the actual feature distance from the query to its fortieth embedding neighbor among the 400 candidates. Its semiaxes equal $d_{40}$ divided by the fitted principal stretches, so the candidates determine only the distance threshold.

\paragraph{Overlap results} Table~\ref{tab:neighbourhoods} reports all three pairwise overlaps (E/G, T/G and E/T) over the retained queries. Of these, only E/T checks directly whether the tensor selects the actual embedding neighbors, which cannot be established from their separate overlaps with geographic neighbors.

Of the 34 primary comparisons with different embedding and tensor selections, 32 share 39 neighbors, one (GeoCLIP at $(0^\circ,0^\circ)$) shares 38, and one (distilled SINR at $(60^\circ\mathrm{N},0^\circ)$) shares 35. Across the twenty queries, the number of the fifteen checkpoint variants with identical selections ranges from 10 to 15. To examine whether agreement differs between land and ocean, we classify the queries with a 30-arcsecond land mask, which replaces the 0.5$^\circ$ mask of Section~\ref{supp:sampling} because the queries are single points, and nine of the twenty queries lie over the ocean. Identical selections occur in 143 of 164 land comparisons and 122 of 135 ocean comparisons, and geographic distance gives identical selections in 3 and 2, respectively, so agreement is similar over land (87\%) and ocean (90\%).

\begin{table}[htbp]\centering\small
\caption{Three-way neighborhood comparison. Entries are median overlaps for $k=40$ among 400 candidates within 1 km; tensors use independent 0.3 km probes. E/G compares actual embedding neighbors with geographic neighbors, T/G compares tensor-predicted neighbors with geographic neighbors, and E/T compares embedding and tensor selections directly. The same valid queries enter all three comparisons.}
\label{tab:neighbourhoods}
\begin{tabular}{@{}lrrrr@{}}\toprule
Checkpoint&Queries&E/G&T/G&E/T\\\midrule
GeoCLIP&20&0.788&0.788&0.975\\
SatCLIP ViT-L10&20&0.762&0.762&1.000\\
SatCLIP R18-L10&20&0.788&0.788&1.000\\
SatCLIP R50-L10&20&0.775&0.775&1.000\\
SatCLIP ViT-L40&20&0.875&0.875&1.000\\
SatCLIP R18-L40&20&0.825&0.825&1.000\\
SatCLIP R50-L40&20&0.875&0.875&1.000\\
SatCLIP ViT-L10 input-rescaled&20&0.900&0.900&1.000\\
SatCLIP ViT-L40 input-rescaled&20&0.850&0.863&1.000\\
SINR cap-1000&20&0.637&0.637&1.000\\
SINR cap-10&20&0.575&0.575&1.000\\
SINR distilled&19&0.575&0.575&1.000\\
CSP-iNat&20&0.713&0.713&1.000\\
CSP-fMoW&20&0.850&0.850&1.000\\
Climplicit&20&0.788&0.788&1.000\\
\bottomrule\end{tabular}
\end{table}

\paragraph{Alternative predictors and larger neighborhoods} To test whether the agreement depends on the fitted geometry rather than on any smooth ranking, Table~\ref{tab:neighbournulls} replaces the fitted tensor with four alternatives, namely geographic distance (equivalent to an isotropic tensor), the fitted tensor with its principal axis rotated by $90^\circ$, the fitted anisotropy with a random principal bearing, and the full fitted tensor of another query of the same checkpoint. In addition, Table~\ref{tab:neighbourextrap} varies the candidate disc and $k$ while keeping the 0.3\,km tensors fixed, so that larger discs test the tensor at several times its fitting radius.

\begin{table}[htbp]\centering\small
\caption{Alternative predictors for the primary neighborhood comparison ($k=40$ of 400 candidates within 1\,km; tensors from separate 0.3\,km probes; 299 valid query-model comparisons). Each predictor ranks the same candidates; entries compare its 40 selections with the 40 actual embedding neighbors.}
\label{tab:neighbournulls}
\begin{tabular}{@{}lrrr@{}}\toprule
Predictor&Identical sets&Median overlap&Mean overlap\\\midrule
Fitted tensor&265&1.000&0.997\\
Isotropic (geographic distance)&5&0.800&0.780\\
Principal axis rotated $90^\circ$&0&0.650&0.620\\
Random principal bearing$^a$&12.3&0.759&0.733\\
Tensor of another query$^b$&2.1&0.776&0.736\\
\bottomrule\end{tabular}
\par\smallskip\parbox{.97\textwidth}{\footnotesize $^a$Fitted anisotropy with a uniformly random bearing; entries average 200 draws per comparison, and the identical-set count is the expected number. $^b$Full fitted tensor of each other valid query of the same checkpoint, averaged over those queries. Query-cluster bootstrap 95\% interval for the fitted-tensor identical fraction: 0.850--0.919.}
\end{table}

\begin{table}[htbp]\centering\small
\caption{Neighborhood agreement beyond the fitting scale. Candidate radius sets the disc from which 400 candidates are drawn; $\tilde d_k$ is the median, over comparisons, of the largest geographic distance from the query among the $k$ embedding neighbors. Tensors are fitted from 0.3\,km probes throughout. Entries give identical-set counts and median overlap with the actual embedding neighbors.}
\label{tab:neighbourextrap}
\begin{tabular}{@{}rrrrrrr@{}}\toprule
Disc radius (km)&$k$&$\tilde d_k$ (km)&\multicolumn{2}{c}{Fitted tensor}&\multicolumn{2}{c}{Isotropic}\\
&&&Identical&Median&Identical&Median\\\midrule
0.3&40&0.12&255/299&1.000&5/299&0.800\\
1&10&0.19&285/299&1.000&50/299&0.800\\
1&20&0.27&277/299&1.000&16/299&0.850\\
1&40&0.40&265/299&1.000&5/299&0.800\\
1&80&0.57&240/299&1.000&0/299&0.825\\
3&40&1.19&245/299&1.000&4/299&0.800\\
6&40&2.42&225/299&1.000&5/299&0.800\\
\bottomrule\end{tabular}
\end{table}

\FloatBarrier

\end{document}